\documentclass[lettersize,journal]{IEEEtran}
\usepackage{multirow}
\usepackage{amsmath,amsfonts}
\usepackage{algorithmic}
\usepackage{algorithm}
\usepackage{array}
\usepackage{textcomp}
\usepackage{stfloats}
\usepackage{url}
\usepackage{verbatim}
\usepackage{graphicx}
\usepackage{cite}
\usepackage{subfigure}
\usepackage{makecell}
\usepackage[table,xcdraw]{xcolor}
\usepackage{float}
\usepackage{color,xcolor}
\usepackage{fancyhdr}
\usepackage{booktabs}
\usepackage{pifont}      
\usepackage{bbding}      
\usepackage{fontawesome}
\begin{document}

\title{ImLiDAR: Cross-Sensor Dynamic Message Propagation Network for 3D Object Detection}

\author{Yiyang Shen, Rongwei Yu, Peng Wu, Haoran Xie, \textit{Senior Member, IEEE}, Lina Gong, Jing Qin, and Mingqiang Wei, \textit{Senior Member, IEEE}
        % <-this % stops a space
\thanks{Y. Shen and R. Yu are with the Key Laboratory of Aerospace Information Security and Trusted Computing, 
Ministry of Education, School of Cyber Science and Engineering, Wuhan 
University, WuHan, China (e-mail:  shenyiyang114@gmail.com, roewe.yu@whu.edu.cn).}
\thanks{P. Wu, L. Gong and M. Wei are with the School
of Computer Science and Technology, Nanjing University of Aeronautics
and Astronautics, Nanjing, China (e-mail: wupengupon1@gmail.com;
gonglina@nuaa.edu.cn; mingqiang.wei@gmail.com).}
\thanks{H. Xie is with the Department of Computing and Decision Sciences, Lingnan University, Hong Kong, China (e-mail: hrxie@ln.edu.hk).}
\thanks{J. Qin is with the School of Nursing, The Polytechnic University of Hong Kong, Hong Kong, China (e-mail: harry.qin@polyu.edu.hk).}
}

\markboth{Journal of \LaTeX\ Class Files,~Vol.~14, No.~8, August~2021}%
{Shell \MakeLowercase{\textit{et al.}}: A Sample Article Using IEEEtran.cls for IEEE Journals}


\maketitle

\begin{abstract}
LiDAR and camera, as two different sensors, supply geometric (point clouds) and semantic (RGB images) information of 3D scenes. However, it is still challenging for existing methods to fuse data from the two cross sensors, making them complementary for quality 3D object detection (3OD).
%, due to large inter-modal discrepancies between different
%representations of images and point clouds, and the introduction of interfering information during the fusion procedure.
%
%potentially provide complementary information for 3D object detection. Beyond the point cloud based wisdom, this paper attempts to exploit the complementary nature of multiple sensors (LiDAR point clouds and camera images) for accurate 3D object detection. 
%
We propose \textbf{ImLiDAR}, a new 3OD paradigm to narrow the cross-sensor discrepancies by progressively fusing the multi-scale features of camera \textbf{Im}ages and \textbf{LiDAR} point clouds.   
%ImLiDAR, which focuses on multi-scale progressive integration of image and LiDAR features to tackle the large inter-s discrepancies. 
ImLiDAR enables to provide the detection head with cross-sensor yet robustly fused features.
To achieve this, two core designs exist in ImLiDAR. First, we propose a cross-sensor dynamic message propagation module to combine the best of the multi-scale image and point features. Second, we raise a direct set prediction problem that allows designing an effective set-based detector to tackle the inconsistency of the classification and localization confidences, and the sensitivity of hand-tuned hyperparameters. 
Besides, the novel set-based detector can be detachable and easily integrated into various detection networks.  Comparisons on both the KITTI and SUN-RGBD datasets show clear visual and numerical improvements of our ImLiDAR over twenty-three state-of-the-art 3OD methods.

% utilizes dynamic message propagation mechanism to tackle the issue mentioned above.
% % To address this issue, we propose a cross-sensor dynamic message propagation network to fuse images and LiDAR point clouds for 3OD, dubbed ImLiDAR. 
% %
% Especially, ImLiDAR introduces a novel module, called cross-sensor dynamic message propagation, which
% % (CDMP), to combine the best of point features and image features in multiple scales.
% %
% % Concretely, CDMP 
% provides an effective manner to combine the best of the point features and image features in multiple scales, since (1) it builds a fine-grained point-wise correspondence between LiDAR and camera image data; (2) it gathers key features while evading harmful information for fusion; (3) it generates essential clues to propagate the semantic image features to enhance the point features; and (4) it does not require BEV data generation procedure which leads to information loss inevitably.
% %
% Further, we design a set-based detector to address two problems arisen by non-maximum suppression (NMS) procedure: (1) the inconsistency of the classification and localization confidence; (2) the sensitivity to hand-tuned hyperparameters, e.g., the predefined IoU threshold.
%
% Comparisons on the KITTI and SUN-RGBD datasets show clear visual and numerical improvements of our ImLiDAR over twenty-three state-of-the-arts.
\end{abstract}

\begin{IEEEkeywords}
ImLiDAR, 3D object detection, Cross sensors, Dynamic message propagation, Set-based detector.
\end{IEEEkeywords}

\section{Introduction}
With the rapid development of autonomous driving, profound progress has been made in 3D object detection from  monocular images \cite{chen2016monocular,wang2021depth,chabot2017deep}, stereo cameras \cite{chen20173d,li2019stereo,chen20153d} and LiDAR point clouds \cite{luo2018fast,yang2018pixor,zhou2018voxelnet}.
% in 3D object detection via different types of sensors (e.g., monocular images \cite{chen2016monocular,wang2021depth,chabot2017deep}, stereo cameras \cite{chen20173d,li2019stereo,chen20153d}, and LiDAR point clouds \cite{luo2018fast,yang2018pixor,zhou2018voxelnet}). 
Among these sensors, LiDAR provides depth and geometric structure information, but its sparsity is causing degraded performance on small-object and long-range perception. Camera images usually possess richer color and semantic information to perceive objects while they lack the depth information for accurate 3D localization. This provides an intriguing and practical question of how to present effective fusion of camera images and LiDAR point clouds for quality 3D object detection.

% Among these sensors, LiDAR is the most commonly used sensor in 3D object detection methods, which provides precise depth and geometric structure information, but discards the semantic information while suffering from the sparsity and disorder of points. Comparatively, camera images usually possess richer semantic information to perceive and distinguish objects while relying heavily on the accuracy of the depth map. This provides the intriguing research opportunity of how to present an effective paradigm to fuse cross-sensor features for more accurate 3D object detection.

\begin{figure*}[!ht] 
\centering
\subfigure[Image]{
\begin{minipage}[b]{0.18\linewidth}
\includegraphics[width=1\linewidth]{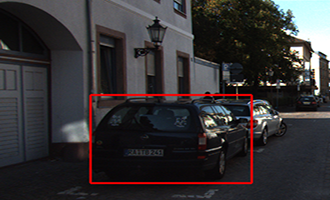}
\includegraphics[width=1\linewidth]{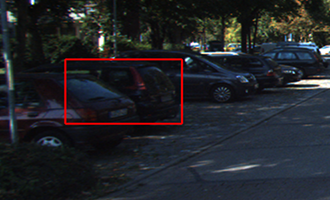}
\end{minipage}
%\caption{fig1}
}\hspace{-2mm}
\subfigure[F-Pointnet \cite{qi2018frustum}]{
\begin{minipage}[b]{0.18\linewidth}
\includegraphics[width=1\linewidth]{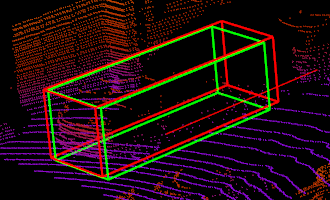}
\includegraphics[width=1\linewidth]{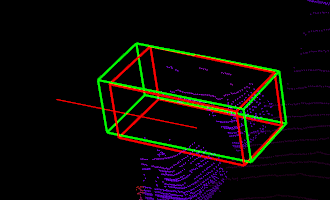}
\end{minipage}
}\hspace{-2mm}
\subfigure[MMF \cite{liang2019multi}]{
\begin{minipage}[b]{0.18\linewidth}
\includegraphics[width=1\linewidth]{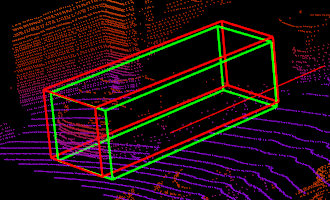}
\includegraphics[width=1\linewidth]{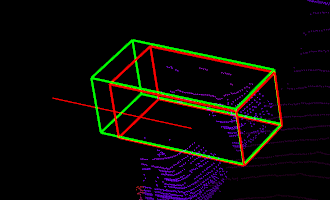}
\end{minipage}
}\hspace{-2mm}
\subfigure[EPNet \cite{huang2020epnet}]{
\begin{minipage}[b]{0.18\linewidth}
\includegraphics[width=1\linewidth]{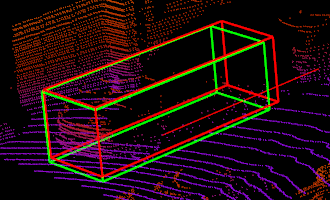}
\includegraphics[width=1\linewidth]{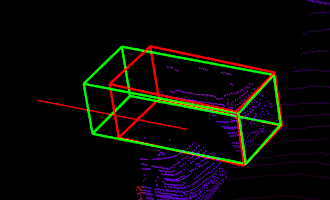}
\end{minipage}
}\hspace{-2mm}
\subfigure[Ours]{
\begin{minipage}[b]{0.18\linewidth}
\includegraphics[width=1\linewidth]{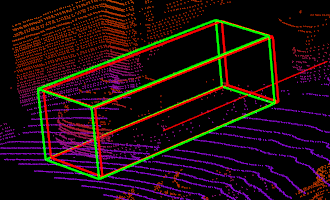}
\includegraphics[width=1\linewidth]{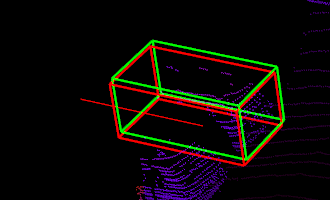}
\end{minipage}
}\hspace{-2mm}
\caption{Visualization results by different categories of fusion methods, i.e., the image-level, BEV-level and feature-level methods.
For the real outdoor scenes where the underexposed instance (row 1) and the occluded instance (row 2) exist, ImLiDAR shows clear visual improvements over the three prevailing cross-sensor methods.  From (a) to (f): (a) the camera image, and (b-e) the 3OD results of F-Pointnet, MMF, EPNet and our ImLiDAR. \textcolor{red}{Red} box and \textcolor{green}{green} box denote GT and the predicted bounding box, respectively.}
\label{fig:intro_result}
\end{figure*}

Recent years have witnessed considerable efforts of information fusion from cross sensors. However, it is still non-trivial to fuse the representations of camera images and LiDAR point clouds, due to their extremely different data characteristics. 
According to different ways of fusion, existing methods are divided into three categories. (1) Image-level methods adopt a cascade strategy by exploiting image annotations to fuse camera images and point clouds in different stages \cite{qi2018frustum,zhao20193d,xu2018pointfusion}. (2) BEV-level methods jointly reason over camera images and the generated BEV data from point clouds \cite{liang2019multi,liang2018deep,chen2017multi,ku2018joint}. (3) Feature-level methods attempt to directly fuse camera images and point clouds by sharing the extracted features between 2D and 3D networks \cite{huang2020epnet,tan2021mbdf,huang2021joint,piergiovanni20214d,liu2021epnet++,yoo20203d}.
Among the three categories, the image-level methods require image annotations, i.e., 2D bounding
boxes, and their performance is easily restricted by each single stage. And the BEV-level methods require generating the BEV data via perspective projection and voxelization; as a result, they usually establish a relatively coarse correspondence between the image and voxel features, and suffer from the loss of 3D information when converting point clouds into the  BEV data.

A recent trend in cross-sensor 3D object detection, which we call feature-level fusion, is to directly fuse the image and point features extracted by 2D and 3D networks. However, when encountering challenging (hard) cases such as underexposed and occluded instances, the performance of existing feature-level fusion methods is still far from satisfactory, due to two major problems.
First, their straightforward fusion strategies assign no weights or coarse weights learned within limited receptive fields to different features. During fusion, there are no crucial clues to keep the original geometric structure without the information loss and avoid introducing new interfering information.
% where lack of crucial clues to keep the original geometric structure without information loss while avoiding introducing new interfering information.
% highlight essential contexts. 
%
% We observe that learned semantic image features inevitably lost essential contextual information while introducing new interfering information in real scenes with abnormal factors, including illumination, occlusion, etc.
%
Second, they heavily rely on the post-processing step of non-maximum suppression (NMS) to remove redundant and near-duplicate results, leading to the inconsistency of the classification confidence and localization confidence. Moreover, NMS requires multiple hand-tuned hyperparameters. For example, a lower threshold of the predefined IoU misses highly overlapped objects while a higher one introduces more false positives.

% First, their straightforward fusion manners inevitably lost essential geometric features while introducing new interfering semantic information during the fusion procedure, and there lacks clues for comprehensively fusing the image and point features. 
%
% Second, these approaches still rely on an additional Non-maximum suppression (NMS) post-processing to remove produced redundant and near-duplicate results, which leads to the inconsistency of the classification confidence and localization confidence and the sensitivity to multiple hand-tuned hyperparameters.
%

To address the above issues, we present a novel cross-sensor dynamic message propagation network, dubbed ImLiDAR, which contains two novel designs for quality 3D object detection. First, we propose a cross-sensor dynamic message propagation (CDMP) module. CDMP targets effective and efficient  fusion of 
camera images and LiDAR point clouds with two key dynamic properties. They are dynamically sampling feature nodes for capturing rich geometric information and filtering harmful semantic information from data of two cross sensors, and dynamically predicting filter weights and affinity matrices as clues for propagating useful image features to enrich the original point features.
% CDMP possesses large receptive fields to capture rich global context information from data of two cross sensors. Subsequently, CDMP predicts image-dependent filter weights and affinity matrices as clues for propagating useful semantic information to enrich the point features.
%
% we project the LiDAR points onto the 2D camera image to build a finer point-wise correspondence. The bilinear interpolation is used to get point-wise image features at the continuous coordinates. 
% %
% Then we regard each element of the point feature and corresponding image feature as a node in graphs, and dynamically sample the neighborhood of a node from the feature graph. This operation can gather the key context features of point clouds and images while releasing harmful information. 
% %
% Based on sampled nodes from point-wise image features, we further predict image-dependent filter weights and affinity matrices as clues for propagating semantic information to enrich the point features. 
%
Second, we formulate a direct set prediction problem and accordingly design a set-based detector to select high-quality 3D bounding boxes with both high classification and localization confidence. Such a set-based detector can avoid the post-processing of NMS, and it can be easily implemented in various detection networks.  

Although ImLiDAR does not need additional image annotations, the complex BEV data, and the commonly used NMS post-processing step,
it usually exhibits better performance over all prevailing cross-sensor methods. 
For example, when encountering real outdoor scenes (see Fig. \ref{fig:intro_result}), cutting-edge models suffer from the condition of poor illuminations and heavy occlusions, while ImLiDAR will not. More results, in terms of visual quality and quantitative accuracy, will be found in Section \ref{experiment}.
In summary, our main contributions are three-fold:
\begin{itemize}
    \item We propose a novel cross-sensor 3D object detection paradigm, namely ImLiDAR, with two core designs, i.e., a cross-sensor dynamic message propagation module and a set-based detector. Extensive experiments in both the outdoor and indoor scenes show clear improvements of ImLiDAR over both the LiDAR-based and cross-sensor methods.
	\item We propose a cross-sensor dynamic message propagation module, combining the best of image and point features without the BEV data or 2D bounding boxes.
    \item We propose a set-based detector to guarantee the consistency between the classification confidence and localization confidence and select high-quality proposals without non-maximum suppression.
    % \item Extensive experiments show that the proposed ImLiDAR outperforms both LiDAR-based and cross-sensor methods on two common 3D object detection benchmark datasets, i.e., the KITTI dataset \cite{geiger2012we} and SUN-RGBD dataset \cite{song2015sun}.
\end{itemize}

\section{Related Work}
% In this section, we first review related works that exploit images based, point cloud based ,multi-modal based methods for 3D object detection. Besides, we introduce related works about graph neural network.
We first review image-based, point cloud-based and cross-sensor methods for 3D object detection. Subsequently, we introduce the recent advance of graph neural networks.

\subsection{Image-based 3D Object Detector}
Many methods focus on camera images, e.g., monocular \cite{ku2019monocular,liu2019deep,reading2021categorical} and stereo images \cite{li2019stereo,wang2019pseudo,wang2021depth}. They  take RGB
images as input to generate 2D bounding boxes and and estimate the corresponding 3D bounding boxes \cite{chen2016monocular,li2019gs3d,chabot2017deep}.
Another way is to conduct depth estimation and design multi-level fusion methods to fuse image features with the depth maps \cite{2018Multi,wang2021depth,ding2020learning}. Particularly, DSGN \cite{chen2020dsgn} provides a simple and
effective one-stage stereo-based 3D detection pipeline that
jointly estimates the depth and detects 3D objects. However, the performance of the image-based methods is bounded due to the absence of depth information.
% due to the absence of precise depth information, all the above methods have to estimate the distance information in 2D images, which brings much difficulty and errors.
% Camera images always play an important role in providing semantic, context and shape information. Early 3D object detection methods focus on camera based solutions with monocular and stereo images. Chen et al. \cite{chen2016monocular} leverage a CNN-based detector to generate 2D bounding boxes with the feature extracted from images and estimates the corresponding 3D bounding boxes. Xu et al. \cite{2018Multi} designs a multi-level fusion method to combine the feature of images with estimated depth map, thus predicting 3D bounding boxes. Chen \cite{chen2020monopair} takes the relationship of paired samples into consideration, utilizing the spatial constraints for partially-occluded objects to improve the accuracy of bounding box. However, due to the absence of precise depth information, all these methods have to estimate the distance information in 2D images, which brings much difficulty and errors.
\subsection{Point Cloud-based 3D Object Detector}
3D object detection methods usually exploit LiDAR point clouds, which provide spatial geometry information to locate the objects. 
They can be divided into voxel-based \cite{zhou2018voxelnet,lang2019pointpillars,deng2020voxel}, and point-based methods \cite{shi2019pointrcnn,huang2020epnet,qi2019deep,zhang2020pc,pan20213d,wu2022casa}. 
Voxel-based models \cite{song2016deep,zhou2018voxelnet,yan2018second,yu2022siev} group point clouds into regular voxels and employ 3D CNNs to learn voxel features for the generation of 3D bounding boxes. To remove 3D CNN layers, PointPillars \cite{lang2019pointpillars} elongates voxels into pillars that are arrayed in a BEV perspective. Point-based approaches \cite{qi2017pointnet,qi2017pointnet++,shi2019pointrcnn,shi2020pv,li2021lidar,huang2020epnet,xie2020pi} sample a fixed number of points as key points via point set abstraction, and aggregate point features around key points with ball query. Recently, most of point-based methods \cite{shi2019pointrcnn,shi2020pv,li2021lidar,huang2020epnet,xie2020pi} formulate a two-stage detection framework, which consists of a region proposal network (RPN) to predict the foreground points and generate 3D proposals, and a refinement network to refine the coarse bounding boxes from RPN. However, all point cloud-based methods suffer from the sparsity of points.

\subsection{Cross-sensor 3D Object Detector}
In realistic self-driving situations, it is insufficient to perform object detection through single types of sensors. Thus, many cross-sensor techniques are proposed to alleviate the shortcomings of single-sensor data.
Current studies are categorized into three groups based on different ways of fusion.

\textbf{Image-level fusion.} Image-level approaches usually exploit camera images in the first stage and reason in LiDAR point
clouds only at the second stage \cite{qi2018frustum,zhao20193d,xu2018pointfusion,du2018general}. 
F-PointNet \cite{qi2018frustum} projects 2D detection results
to 3D space to generate 3D frustums and then adopt PointNet \cite{qi2017pointnet} to regress corresponding 3D boxes from the frustums. V2-SENet \cite{zhao20193d} focuses on utilizing the front view images and frustum point clouds to generate 3D detection results. For these wisdom, the overall performance is bounded by each stage, since they still depend on single sensors. 

\textbf{BEV-level fusion.} MV3D \cite{chen2017multi} is the pioneering attempt to fuse the bird’s eye view (BEV) and front view (FV) representations of cross-sensor data. The follow-up BEV-level methods \cite{ku2018joint,liang2019multi,liang2018deep,chen2017multi} remove the FV branch and only reason over the BEV data and camera images. 
Confuse \cite{liang2018deep} designs a continuous fusion layer to achieve the voxel-wise alignment between the BEV and image feature maps. However, the complex BEV data generation inevitably causes computation costs and the information loss.

\textbf{Feature-level fusion.} A new fashion trend is to fuse each point with the corresponding image pixel instead of fusing the BEV data and camera images \cite{huang2020epnet,tan2021mbdf,huang2021joint,piergiovanni20214d,liu2021epnet++,yoo20203d}.  For example, EPNet \cite{huang2020epnet} designs an end-to-end framework with LiDAR-guided image fusion modules, which assign coarse weights to image features to guide the feature fusion. Similarly, attention fusion modules \cite{tan2021mbdf} and gated fusion modules \cite{yoo20203d} are developed to produce fused features. These methods do not require the generation of 2D bounding boxes and complex BEV, but their fusion manners cannot fully exploit the complementary information of LiDAR point clouds and camera images.

Please note that the proposed ImLiDAR is different from all the above cross-sensor approaches largely, since it combines the best of multi-scale features of camera images and LiDAR point clouds, and does not require any post-processing step of NMS. 

\subsection{Graph Neural Networks}
Graph neural network \cite{scarselli2008graph} have exhibited its powerful ability in many vision tasks, because of their robust capacity of non-local feature aggregation. However, these local-connected graphs can only capture partial long-range contextual information needed for complex vision tasks such as segmentation \cite{landrieu2018large,qi20173d,ma2021fast} and detection \cite{zarzar2019pointrgcn,Shi_2020_CVPR,zhang2020pc,tian2021relation}.
Differently, Zhang et al. \cite{zhang2020dynamic} propose an efficient dynamic graph learning model based on the message propagation mechanism to solve this problem. In this work, we also design a cross-sensor dynamic message propagation (CDMP) module to effectively fuse the LiDAR point features with the corresponding image features, resulting in more comprehensive and discriminative feature representations.

% A few work exploit the fusion of different modal data to obtain rich semantic, context, shape and geometry feature to achieve precise detection. MV3D \cite{chen2017multi} takes a multi-view(such as bird’s eye view and front view) representation of 3D point cloud and an image as input to refine the detection box. F-PointNet \cite{qi2018frustum} utilizes 2D detector to generate 2D bounding box and and then leverages PointNet++ to conduct instance segmentation to produce 3D box. EPNet \cite{huang2020epnet} is the first fusion method that combine each point with corresponding image pixel without generating BEV map,thus formulating a point-pixel feature fused detector. Though fusion method can provide extra image feature to help 3D detector, it may bring possible interference information which is not conducive to feature extraction. Consequently, in this paper, we argue how to combine point cloud and image effectively.
% \subsection{Graph Neural Network}
% In the past several years, Graph neural network\cite{kipf2016semi} has shown its powerful ability in vision tasks. Point-GNN\cite{Shi_2020_CVPR} encodes points in a fixed radius and then constructs graph model to extract feature for 3D object classification and shape feature learning. To comprehensively capture the relationships between points, PC-RGNN \cite{zhang2020pc} proposes a multi-scale graph based module for point cloud completion to help detect 3D object. All these methods denotes that graph neural network is a practical tool in feature learning.
\begin{figure*}[!ht] \centering
	\includegraphics[width=1\linewidth]{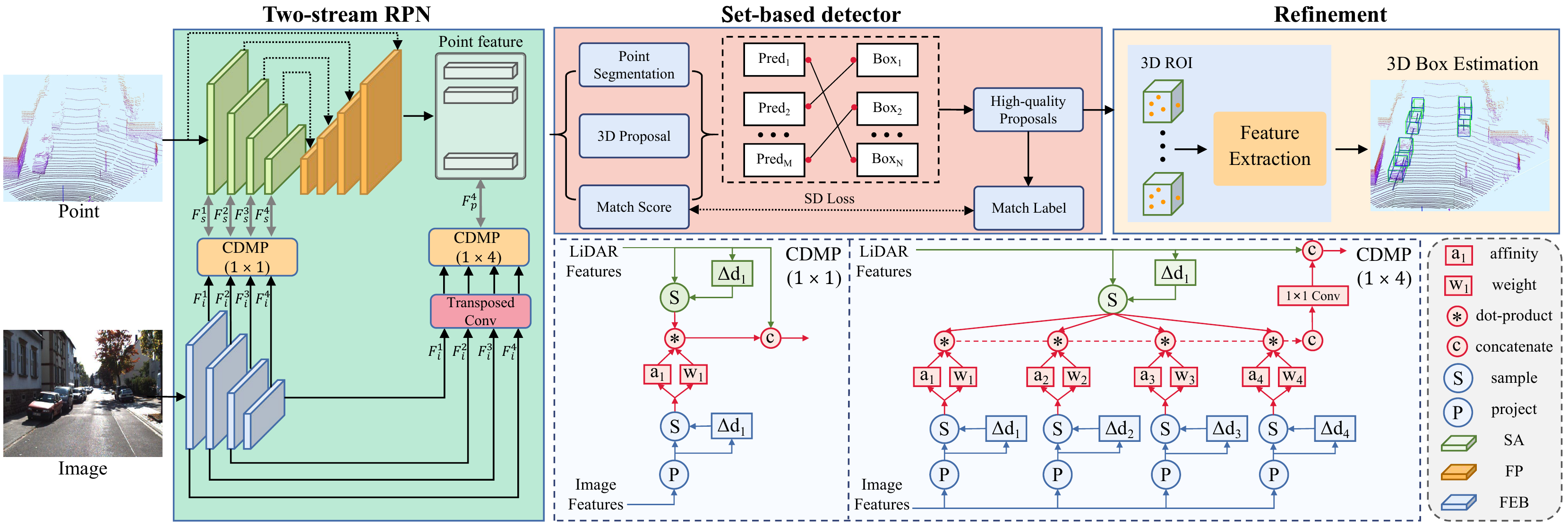}
	\caption{The pipeline of our ImLiDAR. ImLiDAR consists of three cascaded branches, i.e., the two-stream RPN, the set-based detector, and the refinement network. Concretely, the two-stream RPN contains an image stream for extracting image features, a point stream for extracting point features, and well-designed CDMP modules to fuse the geometric point features and semantic image features for enhancing feature representations. Then the set-based detector attempts to select high-quality 3D proposals without the NMS post-processing, and feeds them into the refinement network for further box refinement, leading to more precise 3D object detection results.}
	\label{fig:network framework}
\end{figure*}

\section{Overview}
Cross-sensor fusion has shown its superiority in various applications. 
Primarily, point clouds provide geometric structure information of 3D scenes, and camera images further enrich the point clouds by fulfilling semantic information of the 3D scenes.
To effectively fuse the image and point features in multiple scales for quality 3D object detection, we propose a new 3D object detection paradigm, called ImLiDAR. 
The top level of ImLiDAR, consisting of a two-stream region proposal network (RPN), a set-based detector, and a refinement network, is outlined in Fig. \ref{fig:network framework}.
% is outlined in Fig. \ref{fig:network framework} and consists of a two-stream region proposal network (RPN), a set-based detector, and a refinement network. 

\textbf{Two-stream region proposal network (RPN).} The two-stream RPN consists of a point stream, an image stream, and cross-sensor dynamic message propagation (CDMP) modules. The two-stream RPN 
combines the best of image and point features in multiple scales for 3D object detection, as discussed in Section \ref{RPN}.
% effectively fuse the point features and the image features in multiple scales for quality 3D object detection, as discussed in Section \ref{RPN}.

\textbf{Set-based detector.} Considering that NMS will degrade the detection performance, we newly design a set-based detector to filter out redundant and near-duplicate results to avoid such an NMS step in Section \ref{SED}.

\textbf{Refinement network.} The proposals produced by the set-based detector are fed into the refinement network for further box refinement, leading to more precise 3D object detection results, as discussed in Section \ref{RN}.

\subsection{Preliminary}
Despite the success of graph networks in 2D/3D single-sensor object detection tasks, the attempt to combine advantages from both point clouds and camera images remains scarce. In Section \ref{RPN}, we introduce a cross-sensor dynamic message propagation (CDMP) module to fuse multi-scale features of camera images and LiDAR point clouds. Before going into the details, we will give some basic knowledge of graph message passing used in CDMP.

\textbf{Graph message passing.} Given an input feature map interpreted as the latent feature vectors $H=\{h_{i}\}_{i=1}^{N}$, where $N$ denotes the number of pixels, the goal of the message passing mechanism is to refine the latent feature vectors $H$ by extracting hidden structured information among the feature vectors at different pixel locations.
Therefore, the common message passing network usually converts such feature map into a graph domain by constructing a feature graph $G=\{V,E,A\}$, where $V$ denotes the node set represented by the above latent feature vectors, i.e., $V=\{h_{i}\}_{i=1}^{N}$, $E$ is the edge set, and $A\in R^{N\times N}$ is a binary or learnable matrix with self-loops describing the connections between nodes. 
The common message passing phase, composed of a message calculation step $M^{t}$ and a message updating step $U^{t}$, takes $T$ iterations. 
For the latent feature vector $h_{i}^{(t)}$ at the iteration $t$, it dynamically samples $K$ nodes to connect and form a local field $v_{i} \subset V, v_{i} \in R^{K \times C}, K\ll N$, where $C$ denotes the dimension of the vector. 
The message calculation step for the node $i$ is defined as
\begin{equation}
\begin{aligned}
m_{i}^{(t+1)} &=M^{t}(A_{i,j},\{ h_{1}^{(t)},...,h_{K}^{(t)}\},w_{j})\\
              &=\sum_{j\in \mathcal{N}(i)} A_{i,j}h_{j}^{(t)}w_{j}
\end{aligned}
\end{equation}
where $A_{i,j}$ denotes the connection relationship between latent nodes $h_{i}^{(t)}$ and $h_{j}^{(t)}$, $\mathcal{N}(i)$ represents a self-included neighborhood of the node $h_{i}^{(t)}$, and $w_{j}\in R^{C\times C}$ is a transformation matrix for message calculation on the hidden node $h_{j}^{(t)}$. 
Then the message updating step $U^{t}$ obtains the updated latent feature vector $h_{i}^{(t+1)}$ with a linear combination of the calculated message $m_{i}^{(t+1)}$ and the original feature vector $h_{i}^{(t)}$ at the node position $i$:
\begin{equation}
h_{i}^{(t+1)} =U^{t}(h_{i}^{(t)},m_{i}^{(t+1)})=\sigma(h_{i}^{(t)}+\alpha_{i}^{m}m_{i}^{(t+1)})
\label{eq:original_refine}
\end{equation}
where $\alpha_{i}^{m}$ denotes a learnable parameter to scale the message, and $\sigma(.)$ is a non-linearity function, e.g., ReLU. 
By propagating the message on each node with $T$ steps, the module finally obtains the refined features. 
Especially, we fuse the image and point features via the graph message passing mechanism as:
\begin{equation}
h_{i}^{(t+1)} =\sigma(h_{i}^{(t)}||\alpha_{i}^{m}m_{i}^{(t+1)})
\label{eq:refine}
\end{equation}
where $||$ denotes the channel-wise concatenation operation.

\begin{figure}[!t] \centering
	\includegraphics[width=1\linewidth]{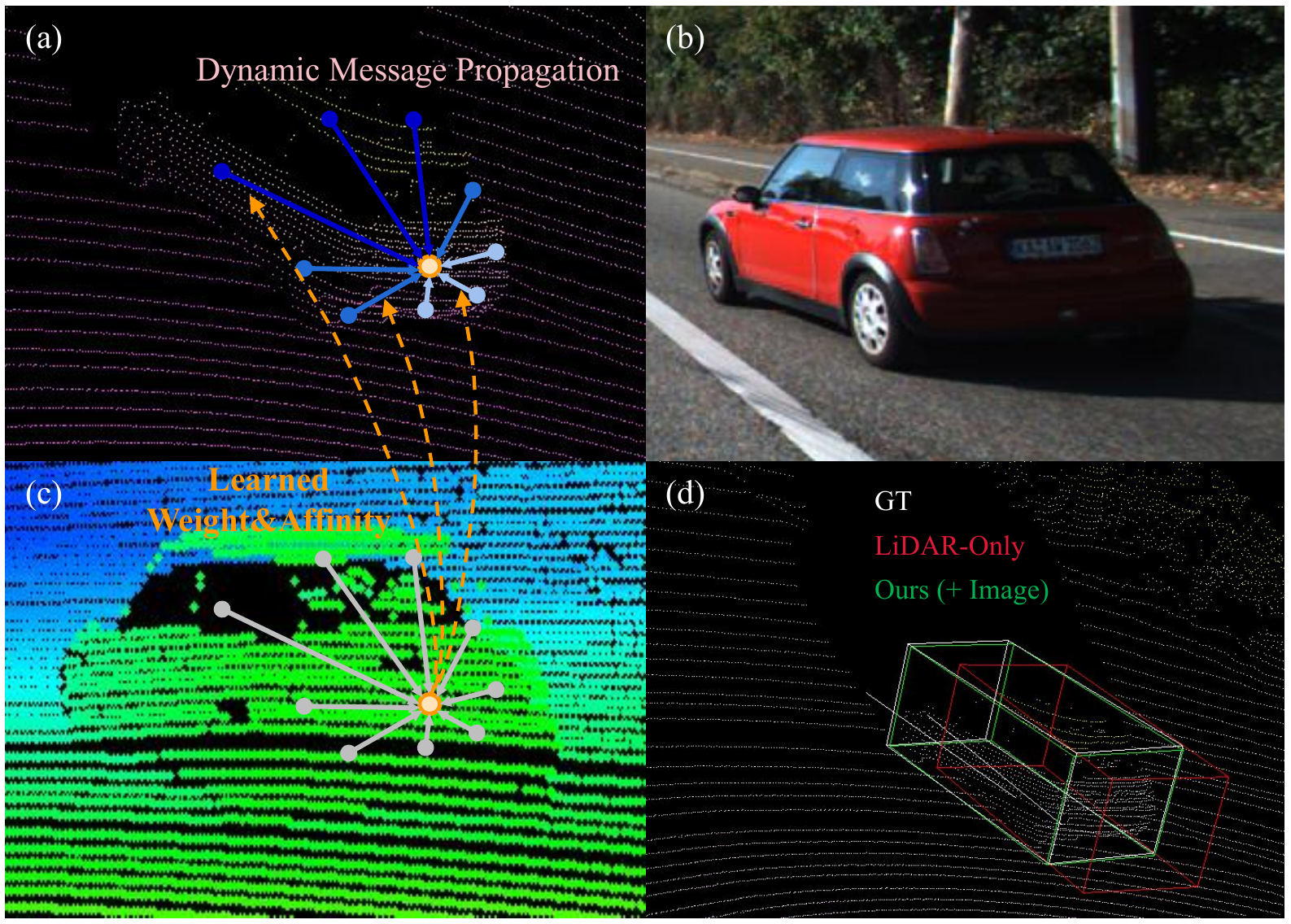}
	\caption{Illustration of the cross-sensor dynamic message propagation (CDMP) module. CDMP first dynamically samples context-aware nodes in the LiDAR point features (a) and point-wise image features (c), which are extracted by projecting the source LiDAR (a) onto the image plane (b). Then it predicts hybrid image-dependent filter weights and affinity matrices as clues for propagating semantic information to enrich the point features. (d) demonstrates that our ImLiDAR can effectively fuse the image and point features, leading to significant improvement of the 3D object detection performance. Note that white, red and green boxes represent the ground truth, predicted bounding boxes of the LiDAR-based detector \cite{shi2019pointrcnn} and ours (LiDAR+Image), respectively.}
	\label{fig:intro}
\end{figure}

\section{ImLiDAR}
ImLiDAR consists of a two-stream RPN, a set-based detector, a refinement network, and the defined loss function. 

\subsection{Two-stream RPN}\label{RPN}
Our two-stream RPN consists of a point stream, an image stream, and cross-sensor dynamic message  propagation (CDMP) modules in Fig. \ref{fig:network framework}. 
The point stream and image stream are designed for extracting multi-scale geometric point features and semantic image features, respectively. 
The CDMP modules are employed to fuse the image and point features in different scales, resulting in more robust and discriminative representations.

\textbf{Image stream.} The image stream, depicted in Fig. \ref{fig:network framework}, takes camera images as input to extract multi-scale semantic image features. 
Concretely, the architecture of the image stream consists of four feature extract blocks (FEBs), which both include two 3×3 convolution layers followed by a batch normalization layer and a ReLU activation function. 
$F_{i}^{k}$ $(k=1,2,3,4)$ denotes the 
multi-scale features extracted from four FEBs, which provide adequate semantic information to enrich the point features in different scales. 
At the end of the image stream, we feed these multi-scale image features into four parallel transposed convolution layers to obtain image features with the same size as the original image, which are used to enrich the final point features, resulting in the generation of more high-quality proposals.

\textbf{Point stream.} For the point stream, we employ PointNet++ \cite{qi2017pointnet++} as our backbone network. 
The point stream takes LiDAR point clouds as input and utilizes four set-abstraction (SA) modules with multi-scale grouping to subsample points into groups with the sizes of 4096, 1024, 256, 64, and four feature propagation (FP) modules to recover the point resolution.
Especially, $F_{s}^{k}$ $(k=1,2,3,4)$ and $F_{p}^{k}$ $ (k=1,2,3,4)$ represent the outputs of SA and FP layers in different scales, respectively.
With the aid of CDMP ($1\times 1$) modules, we can effectively fuse the point features $F_{s}^{k}$ with the image features $F_{i}^{k}$ at different levels.
Further, we apply the CDMP ($1\times 4$) module at the end of the point stream, which enables the point feature vectors $F_{p}^{4}$ to possess different level semantic information from the image features $F_{i}^{k}$ $(k=1,2,3,4)$.
Similar to PointRCNN \cite{shi2019pointrcnn}, given the final point features, we first append a box regression head for 3D proposal generation and a segmentation head for foreground point segmentation.
Moreover, we append an additional match head to estimate the match score for the set-based detector. The match scores mean that the probability of each predicted 3D bounding box is retained by the set-based detector.

\begin{figure}[!t] \centering
	\includegraphics[width=1\linewidth]{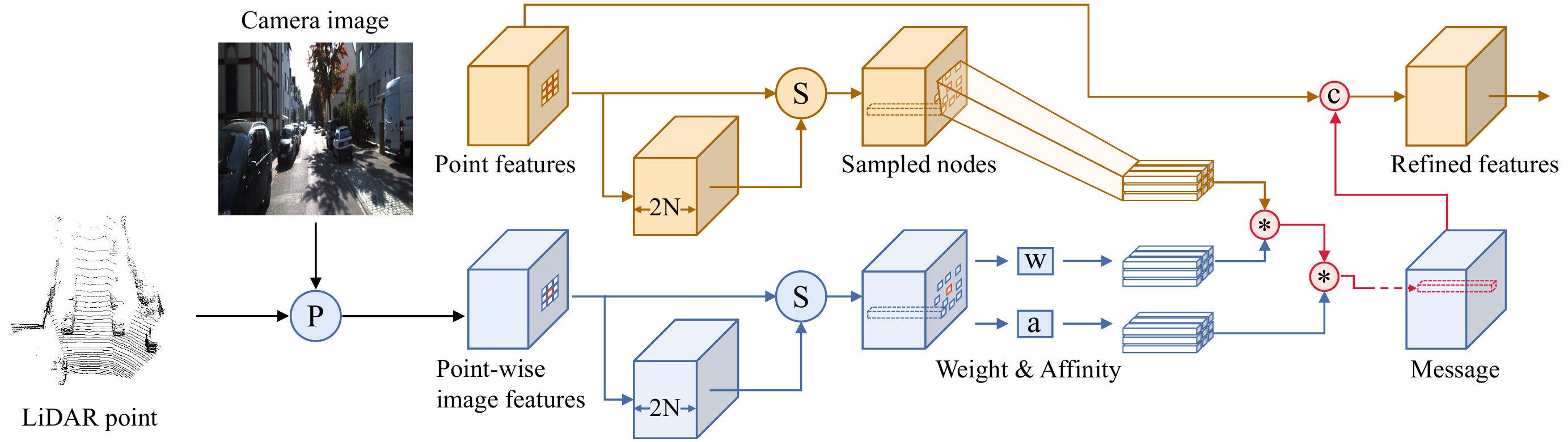}
	\caption{Illustration of CDMP in a single scale pattern. We first project the LiDAR points onto the 2D camera image to obtain the corresponding point-wise image features. Then we sample the dynamic nodes from the image and point feature graphs, and predict the filter weights and affinity matrices from image features to propagate the semantic message.}
	\label{fig:dmmp}
\end{figure}

\textbf{Cross-sensor dynamic message propagation (CDMP).} 
To combine the best of multi-scale image and point features, we design a novel CDMP module. 
The more detailed scheme of CDMP is further depicted in Fig. \ref{fig:intro}. In particular, it includes three steps: (1) generating the fine-grained point-wise correspondence and point-wise image features; (2) dynamically sampling on image and point feature graphs to select the most object-relevant nodes; and (3) dynamically predicting hybrid filter weights and affinity matrices for message propagation.

% , which are regarded as clues to enhance the point features with essential image features.

The CDMP module in a single scale pattern is shown in Fig. \ref{fig:dmmp}. 
First, we project the LiDAR points onto the 2D camera image based on the calibration matrix, which is usually provided by the benchmark datasets, to generate a finer point-wise correspondence between LiDAR points and camera images. 
Concretely, for a particular point $p(x,y,z)$ in the point cloud, we obtain its corresponding position $p'(x,y)$ in the camera image. 
Then we input both the image feature map and sampling position $p'$ into the bilinear interpolation to produce point-wise image features at the continuous coordinates. 
%
% This operation helps to build a fine-grained point-wise correspondence and generate corresponding point-wise image features.

% We first extract corresponding point-wise image features by projecting the LiDAR points onto the 2D camera image based on the calibration matrix, which is usually provided by the benchmark datasets, e.g., the KITTI dataset \cite{geiger2012we}. Then the bilinear interpolation is used to get the image feature at the continuous coordinates. After that, to gather the key image and point features while filtering the harmful semantic information caused by illumination, occlusion, etc, we regard the image feature map and point feature map as two graphs, and adopt deformable convolution \cite{zhang2020dynamic,dai2017deformable} to dynamic sample context-aware nodes in both image and point feature graphs. 
%
Based on the fine-grained point-wise image features, CDMP possesses two novel dynamic properties, i.e., dynamically sampling feature nodes and dynamically predicting filter weights and affinity matrices.
% To gather both essential contextual features while filtering the harmful semantic information caused by illumination, occlusion, etc, 
We regard the image feature map and point feature map as two graphs.
% then adopt deformable convolution \cite{zhang2020dynamic,dai2017deformable} to dynamic sample context-aware nodes in both image and point feature graphs.
%
For each node $v_{i}$ in the point feature node set $V= \{v_{i}{\}}_{1}^{N}$, where $N$ is the total number of pixels, the sampling number $K$ determines its receptive field. 
To adaptively sample relevant nodes for $v_{i}$, we denote $\Delta d_{i,j} \in R^{D}$ as the predicted walk, which makes the module walk around to sample the relevant node $v_{i,j}$ with $j \in \mathcal{N}(i)$, where $\mathcal{N}(i)$ contains $K$ number of sampled nodes for $v_{i}$ and $D = 2$ represents the space dimension along the height and width. Especially, we describe such node walk as a matrix transformation:
\begin{equation}
\Delta d_{i,j} = W_{i,j}h_{i}+b_{i,j}
\label{eq:walk}
\end{equation}
where $W_{i,j}$ and $b_{i,j}$ both are matrix transformation parameters learned on point graph nodes, and $h_{i}$ denotes the latent vector for $v_{i}$. 
Due to the difference between point and image modalities, we generate another walk $\Delta \bar{d}_{i,j}$ for uniformly neighboring nodes on the image feature graph. The dynamic walk $\Delta d_{i,j}$ for each point feature node, as well as $\Delta \bar{d}_{i,j}$ for each image feature node, is generated by applying 3 $\times$ 3 convolution layers according to Equation \ref{eq:walk}. 
Based on the above random walks, we adopt the deformable convolution \cite{dai2017deformable} to obtain dynamic sample nodes $\hat{v}_{i,j}^{l}$ and $\bar{v}_{i,j}^{l}$ from point and image feature graphs at the level $l$, respectively. This dynamically sampling operation enables to efficiently gather long-range context information and only select a subset of the most important feature nodes in the two graphs.

Based on the sampled image feature nodes $\bar{v}_{i,j}^{l}$, we apply 3$\times$3 convolution layers to generate the affinity matrix $A_{i,j}^{l}$ and transformation matrix $w_{i,j}^{l}$, which are used as clues for propagating the useful image features to enhance the point features, which can be formulated as:
% the image feature nodes $\bar{v}_{i,j}^{l}$ learns to further generate affinity matrix $A_{i,j}^{l}$ and transformation matrix $w_{i,j}^{l}$ as:
\begin{equation}
\{A_{i,j}^{l};w_{i,j}^{l}\} = \bar{W}_{i,j}^{l}\bar{v}_{i,j}^{l}+\bar{b}_{i,j}^{l}
\end{equation}
where $\bar{W}_{i,j}^{l}$ and $\bar{b}_{i,j}^{l}$ are matrix transformation parameters generated by dynamic sample image nodes. Then the calculated message is summarized as:
\begin{equation}
\begin{aligned}
m_{i}^{t+1} &= \sum_{l\in L} \sum_{j \in \mathcal{N}(i)} \beta_{l}A_{i,j}^{l}\hat{h}_{j}^{l,(t)}w_{i,j}^{l}\\
            &= \sum_{l\in L} \sum_{j \in \mathcal{N}(i)} \beta_{l}A_{i,j}^{l}\delta(\hat{h}_{j}^{l,(t)}|V;j;\Delta d_{i,j})w_{i,j}^{l}
\end{aligned}
\label{eq:message}
\end{equation}
where $L$ denotes the layer from different level stages, $\hat{h}_{j}^{l,(t)}$ is the latent vector for dynamic point feature nodes $\hat{v}_{i,j}^{l}$ calculated by $\Delta d_{i,j}$ over the whole nodes $V$ of the point graph.
$\beta_{l}$ and $\delta(.)$ represent the balance weight and bilinear sampler, respectively. 
In CDMP ($1\times 4$) module, all messages are calculated as Equation \ref{eq:message} using group convolution layers and concatenated into a 1$\times$1 convolution layer. 
Then the result is concatenated again with the original point features to obtain the final refined point features with semantic image information, as described in Equation \ref{eq:refine}.

\subsection{Set-based Detector}\label{SED}
Most of existing 3D object detectors predict a larger number of bounding boxes than the number of the real objects in the scene. In view of such fact, non-maximum suppression (NMS) is often a necessary post-processing step.

NMS includes two parts: (1) Selecting 3D bounding boxes with the maximum scores after ranking the proposals according to the classification scores. However, NMS may filter out the bounding boxes with low classification scores but large overlaps; this leads to the inconsistency of the classification confidence and localization confidence. (2) Removing any 3D bounding box which possesses an overlap greater than a predefined IoU threshold. It makes the current detectors with a dilemma: a lower threshold leads to missing highly overlapped objects while a higher one introduces more false positives in crowded scenes.
% In view of the fact that existing 3D detection models still heavily rely on a non-maximum suppression (NMS) post-processing to remove redundant and near-duplicate bounding boxes. This brings about two drawbacks: (1) The inconsistency between the classification confidence and localization confidence. Since the classification scores are commonly used as the metric for ranking the proposals, the NMS may filter out the bounding boxes with low classification scores but large overlaps; (2) The sensitivity to multiple hand-tuned hyperparameters, e.g., the predefined IoU threshold. A lower threshold leads to missing highly overlapped objects while a higher one brings in more false positives, especially in a crowded scene. 
% The commonly used NMS usually filters low-quality bounding boxes based on their classification confidence and IoU. The former leads to the inconsistency of the classification confidence and the localization confidence, and the latter makes the current detectors with a dilemma for the hand-tuned threshold of NMS: a lower threshold leads to missing highly overlapped objects while a higher one brings in more false positives.

Motivated by DETR \cite{carion2020end}, we design a set-based detector to address the above problems caused by NMS. 
We design three sets of bounding boxes: the set of predicted 3D bounding boxes $B_{pre}=\{\bar{b}\}^{M}_{1}$ from RPN, the set of GT bounding boxes $B_{gt}=\{b\}^{N}_{1}$, and the set of output bounding boxes $B_{out}=\{\Tilde{b}\}^{N}_{1}$,
where $M$ and $N$ represent the number of bounding boxes, and $M\gg N$. 
The number $N$ is larger than or equal to the number of the real objects in the scene.
% , since the set $B_{gt}$ may contain repeated ground truth bounding boxes. 
% 
To obtain the high-quality bounding box set $B_{out}$, we perform a bipartite graph matching, which is simpler and more effective than NMS. 
We compute the match cost for each predicted box $\bar{b}$ with each ground truth box $b$.

\textbf{Bipartite matching.} We define a match cost for a pair of predicted box $\bar{b}$ and GT box $b$, which is formulated as:
\begin{equation}
C_{match}(\bar{b},b)=-\log(c\times \frac{Area(\bar{b} \cap b)}{Area(\bar{b} \cup b)})
\end{equation}
% where $\bar{b}$ and $b$ represents the predicted 3D bounding box of $B_{pre}$ and the repeated ground truth bounding boxes of $B_{gt}$, and $c$ denotes the classification confidence for $\bar{b}$. 
where $c$ denotes the classification confidence for $\bar{b}$. We compute the optimal bipartite matching between all the predicted boxes $\bar{b}$ and GT boxes $b$ using the Hungarian algorithm \cite{kuhn1955hungarian}.
Therefore, each GT box $b$ is successfully matched with a predicted 3D bounding box $\bar{b}$ with both large overlaps and high classification possibilities. 
We regard these matched predicted 3D bounding boxes as positive samples and the others as negative samples. During the training step, these positive samples constitute the output high-quality bounding boxes $B_{out}$, which are fed into the refinement network. 
We also generate a match label vector $\Tilde{M}\in R^{M \times 1}$ according to the match results. 
Particularly, if the predicted 3D bounding box is a positive sample, its corresponding value of $\Tilde{M}$ will be set to 1, and the others are set to 0. 
Then we adopt the focal loss \cite{lin2017focal} between the match label and match score from the match head of RPN:
\begin{equation}
L_{SD} = -\alpha(1-c_{m})^{\gamma}\log c_{m}
\end{equation}
where $c_{m}$ represents the probability of the predicted box $\bar{b}$ is the positive sample, and $\alpha = 0.25, \gamma = 2$ are kept as in \cite{lin2017focal}. During the testing step, we directly select 3D bounding boxes with the highest match score to constitute $B_{out}$ for the refinement network.

% Then we compute the optimal bipartite matching between all the predicted boxes $\bar{b}$ and ground truth boxes $b$ using the Hungarian algorithm \cite{kuhn1955hungarian}. Towards searching the best match with the least match cost, each ground truth box $b$ will get the match to a predicted 3D bounding box $\bar{b}$ with both large overlaps and high classification possibilities. We regard these matched predicted 3D bounding boxes as positive samples and others as negative samples, and generate a match mask vector, where the labels of positive and negative samples are set to be 1 and 0, respectively.
% Then we adopt the focal loss \cite{lin2017focal} as our classification loss to balance the positive and negative samples, which can be formulated as:
% \begin{equation}
% L_{set-based} = -\alpha(1-c_{m})^{\gamma}\log c_{m}
% \end{equation}
% where $c_{m}$ represent the probability of the proposal belong to set of output high-quality bounding boxes $B_{out}$. In the training process, these positive samples consitute the output high-quality bounding boxes $B_{out}$ and are fed into the refinement network. In the testing process, we just select $K$ proposals with the highest match score for the refinement network.
% In general, the proposed set-based detector effectively selects the set of output high-quality bounding boxes $B_{out}$ without hand-crafted priori (e.g, the threshold of IoU), and is easy to plug into existing 3D detection paradigms.

\subsection{Refinement Network}\label{RN}
We feed the set of proposals $B_{out}$ from the set-based detector into the refinement network to refine the box locations and orientations for final predictions. Similar to PointRCNN \cite{shi2019pointrcnn}, for each input proposal, we randomly select 512 points as its 3D RoI feature descriptor. For those proposals with less than 512 points, the descriptor is padded with zeros. The refinement network is composed of three SA layers for extracting a compact global descriptor for each 3D ROI, and two 1×1 convolution layers as two detection heads for classifying and regressing the final 3D objects.

\subsection{Total Loss Function}
We present the loss functions. We adopt a multi-task loss function for jointly optimizing the two-stream RPN, the set-based detector and the refinement network, which can be defined as:
% For jointly optimizing the two-stream RPN, the set-based detector and the refinement network, we follow the multi-task loss function of PointRCNN \cite{shi2019pointrcnn}, which can be defined as follows:
\begin{equation}
L_{total} = L_{rpn}+L_{rcnn}+\lambda L_{SD}
\end{equation}
\begin{equation}
\{L_{rpn};L_{rcnn}\} = L_{cls}+L_{reg}
\end{equation}
where $L_{rpn}$ and $L_{rcnn}$ represent the training objective for the two-stream RPN and the refinement network. 
They both contain a classification loss and a regression loss. 
Concretely, we adopt the focal loss \cite{lin2017focal} as the classification loss to balance the positive and negative samples as:
\begin{equation}
L_{cls} = -\alpha(1-c_{t})^{\gamma}\log c_{t}
\end{equation}
where $c_{t}$ represents the probability of the point in consideration belonging to the ground truth category. And we keep the default settings $\alpha = 0.25, \gamma = 2$ as suggested by \cite{lin2017focal}.

In the LiDAR coordinate system, a 3D bounding box is represented as $(x,y,z,h,w,l,\theta)$, where $(x,y,z)$ is the object center location, $(h,w,l)$ is the object size, and $\theta$ denotes the object orientation. 
Following PointRCNN \cite{shi2019pointrcnn}, we adopt the bin-based regression loss as our regression loss function to estimate 3D bounding boxes of objects. 
Concretely, we split the neighboring area of each foreground point into several bins. 
The bin-based loss first predicts which bin $\Tilde{b}_{u}$ the center point belongs to, and regresses the residual offset $\Tilde{r}_{u}$ within the bin. Thus, the regression loss is formulated as:
\begin{equation}
L_{reg} = \sum_{u\in x,z,\theta}E(\Tilde{b}_{u},b_{u})+\sum_{u\in x,y,z,h,w,l,\theta}S(\Tilde{r}_{u},r_{u})
\end{equation}
where $E$ and $S$ denote the cross entropy loss and the smooth L1 loss, respectively. $b_u$ and $r_u$ denote the ground truth of the bins and the residual offsets.

% \usepackage{multirow}
% \usepackage{multirow}

% \usepackage{multirow}
\begin{table*}[!t]
\centering
\caption{Quantitative comparisons with state-of-the-art methods on the test set of the KITTI dataset.}
\setlength{\tabcolsep}{1.65mm}{
\begin{tabular}{c|c|cccc|cccc|cccc}
\toprule
                         &                            & \multicolumn{4}{c|}{Car(IoU=0.7)} & \multicolumn{4}{c|}{Pedestrian(IoU=0.5)}                                                                                                                                                                       & \multicolumn{4}{c}{Cyclist(IoU=0.5)}                                                                                                                                                                         \\
\multirow{-2}{*}{Method} & \multirow{-2}{*}{Modality} & Easy  & Moderate & Hard  & mAP   & Easy                                             & Moderate                                         & Hard                                                 & mAP                                              & Easy                                             & Moderate                                         & Hard                                                & mAP                                              \\\midrule
SECOND \cite{yan2018second}                & LiDAR                      & 87.44 & 79.46    & 73.97 & 80.29 & -                                             & -                                             & -                                                 & -                                             & -                                             & -                                             & -                                                & -                                             \\
PointPillars \cite{lang2019pointpillars}            & LiDAR                      & 82.58 & 74.31    & 68.99 & 75.29 & 51.45                                            & 41.92                                            & 38.89                                                & 44.09                                            & 77.10                                             & 58.65                                            & 51.92                                               & 62.56                                            \\
PointRCNN \cite{shi2019pointrcnn}                & LiDAR                      & 86.96 & 75.64    & 70.70  & 77.76 & 47.98 & 39.37 & 36.01     & 41.12 & 74.96                                            & 58.82                                            & 52.53                                               & 62.10                                             \\
STD \cite{yang2019std}                      & LiDAR                      & 87.95 & 79.71    & 75.09 & 80.91 & 53.29 & 42.47 & 38.35     & 44.70  &  78.69 & 61.59 & 55.30 & 65.19 \\
3DSSD \cite{yang20203dssd}                   & LiDAR                      & 88.36 & 79.57    & 74.55 & 80.82 &  54.64 &  44.27 &  40.23     &  46.38 &  82.48 &  64.10  &  56.90     &  67.82 \\
SA-SSD \cite{he2020structure}                  & LiDAR                      & 88.75 & 79.79    & 74.16 & 81.03 & -                                             & -                                             & -                                                 & -                                             & -                                             & -                                             & -                                                & -                                             \\
PV-RCNN \cite{shi2020pv}                 & LiDAR                      & 90.25 & 81.43    & 76.82 & 82.83 &  52.17 &  43.29 &  40.29 &  45.25 &  78.60  &  63.71 &  57.65    &  66.65 \\
MGAF-3DSSD \cite{li2021anchor}              & LiDAR                      & 88.16 & 79.68    & 72.39 & 80.07 &  50.65                     &  43.09                     &  39.65                         &  44.46                     &  80.64                     &  63.43                     &  55.15                        &  66.40                      \\
HVPR \cite{noh2021hvpr}                    & LiDAR                      & 86.38 & 77.92    & 73.04 & 79.11 & -                                             & -                                             & -                                                 & -                                             & -                                             & -                                             & -                                                & -                                             \\
CIA-SSD \cite{zheng2021cia}                 & LiDAR                      & 89.59 & 80.28    & 72.87 & 80.91 & -                                             & -                                             & -                                                 & -                                             & -                                             & -                                             & -                                                & -                                             \\
CT3D \cite{sheng2021improving}                     & LiDAR                      & 87.83 & 81.77    & 77.16 & 82.25 & -                                             & -                                             & -                                                 & -                                             & -                                             & -                                             & -                                                & -                                             \\
SASA \cite{chen2022sasa}                    & LiDAR                      & 88.76 & 82.16    & 77.16 & 82.69 & -                                             & -                                             & -                                                 & -                                             & -                                             & -                                             & -                                                & -                                             \\
SVGA-Net \cite{he2020svga}                 & LiDAR                      & 87.33 & 80.47    & 75.91 & 81.23 &  48.48                     &  40.39                     &  37.92                         &  42.26                     &  78.58                     &  62.28                     &  54.88                        &  65.24                     \\\bottomrule
MV3D \cite{chen2017multi}                   & LiDAR + RGB                & 74.97 & 63.63    & 54.00    & 64.20  & -                                             & -                                             & -                                                 & -                                             & -                                             & -                                             & -                                                & -                                             \\
Confuse \cite{liang2018deep}                 & LiDAR + RGB                & 83.68 & 68.78    & 61.67 & 71.38 & -                                             & -                                             & -                                                 & -                                             & -                                             & -                                             & -                                                & -                                             \\
F-Pointnet \cite{qi2018frustum}              & LiDAR + RGB                & 82.19 & 69.79    & 60.59 & 70.86 &  50.53 &  42.15 &  38.08     &  43.59 &  72.27 &  56.12 &  49.01    &  59.13 \\
MMF \cite{liang2019multi}                     & LiDAR + RGB                & 88.40  & 77.43    & 70.22 & 78.68 & -                                             & -                                             & -                                                 & -                                             & -                                             & -                                             & -                                                & -                                             \\
3D-CVF \cite{yoo20203d}                  & LiDAR + RGB                & 89.20  & 80.05    & 73.11 & 80.79 & -                                             & -                                             & -                                                 & -                                             & -                                             & -                                             & -                                                & -                                             \\
PointPainting \cite{vora2020pointpainting}           & LiDAR + RGB                & 82.11 & 71.70     & 67.08 & 73.63 &  50.32 &  40.97 & 37.84                                                &  43.05 &  77.63 &  63.78 &  55.89    &  65.77 \\
EPNet \cite{huang2020epnet}                   & LiDAR + RGB                & 89.81 & 79.28    & 74.59 & 81.23 & -                                             & -                                             & -                                                 & -                                             & -                                             & -                                             & -                                                & -                                             \\
Fast-CLOCs \cite{pang2022fast}              & LiDAR + RGB                & 89.10  & 80.35    & 76.99 & 82.14 &  52.10  &  42.72 &  39.08     &  44.63 &  82.83 &  65.31 &  57.43    &  68.53 \\
Focals Conv \cite{chen2022focal}              & LiDAR + RGB                & 90.55  & 82.28    & 77.59 & 83.47 &  -  &  - &  -    &  - &  - &  - &  -    &  - \\
CAT-Det \cite{2204.00325}                 & LiDAR + RGB                & 89.87 & 81.32    & 76.68 & 82.62 &  54.26 &  45.44 &  41.94 &  47.21 &  83.68 &  68.81 &  61.45    &  71.31 \\\bottomrule
ImLiDAR                  & LiDAR + RGB                & 90.98 & 83.23    & 77.67 & 83.96 &  55.38 &  46.26 &  42.38     &  48.01 &  84.22 &  68.89 &  61.80     &  71.63\\\bottomrule
\end{tabular}
}
\label{compare}
\end{table*}

\begin{table*}[]
\centering
\caption{Quantitative comparisons with the state-of-the-art methods on the test set of the SUN-RGBD dataset.}
\begin{tabular}{ccccccccccccc}
\toprule
Method      & Modality    & bathtub       & bed           & bookshelf     & chair         & desk          & dresser       & nightstand    & sofa          & table         & toilet        & mAP           \\\midrule
VoteNet \cite{qi2019deep}& LiDAR &74.4 &83.0 &28.8& 75.3 &22.0 &29.8& 62.2 &64.0 &47.3& 90.1& 57.7\\

MLCVNet \cite{xie2020mlcvnet}    & LiDAR       & 79.2          & 85.8          & 31.9          & 75.8          & 26.5          & 31.3          & 61.5          & 66.3          & 50.4          & 89.1          & 59.8          \\
H3DNet \cite{zhang2020h3dnet}    & LiDAR       & 73.8          & 85.6          & 31.0          & 76.7          & 29.6          & 33.4          & 65.5          & 66.5          & 50.8          & 88.2          & 60.1          \\
HGNet \cite{chen2020hierarchical}      & LiDAR       & 78.0          & 84.5          & 35.7          & 75.2          & 34.3 & 37.6 & 61.7          & 65.7          & 51.6          & 91.1          & 61.6          \\
MLCVNet++ \cite{xie2021vote}& LiDAR &79.3&85.3&36.5&77.1&28.7&31.6&61.4&68.3&50.7&90.0&60.9\\
Group-Free-3D \cite{liu2021group}&LiDAR&80.0 &\textbf{87.8} &32.5&\textbf{79.4}&32.6&36.0&\textbf{66.7}&\textbf{70.0}&53.8&91.1&63.0 \\
Pointformer \cite{pan20213d}& LiDAR       & 80.1 & 84.3          & 32.0          & 76.2          & 27.0          & 37.4          & 64.0          & 64.9          & 51.5          & \textbf{92.2}          & 61.1          \\\bottomrule
DSS \cite{song2016deep}        & LiDAR + RGB & 44.2          & 78.8          & 11.9          & 61.2          & 20.5          & 6.4           & 15.4          & 53.5          & 50.3          & 78.9          & 42.1          \\
2D-driven \cite{lahoud20172d}   & LiDAR + RGB & 43.5          & 64.5          & 31.4          & 48.3          & 27.9          & 25.9          & 41.9          & 50.4          & 37.0          & 80.4          & 45.1          \\
COG \cite{ren2016three}        & LiDAR + RGB & 58.3          & 63.7          & 31.8          & 62.2          & 45.2          & 15.5          & 27.4          & 51.0          & 51.3          & 70.1          & 47.6          \\
PointFusion \cite{xu2018pointfusion}& LiDAR + RGB & 37.3          & 68.6          & \textbf{37.7} & 55.1          & 17.2          & 24.0          & 32.3          & 53.8          & 31.0          & 83.8          & 44.1          \\
F-PointNet \cite{qi2018frustum}  & LiDAR + RGB & 43.3          & 81.1          & 33.3          & 64.2          & 24.7          & 32.0          & 58.1          & 61.1          & 51.1          & 90.9 & 54.0          \\
EPNet \cite{huang2020epnet}      & LiDAR + RGB & 75.4          & 85.2          & 35.4          & 75.0          & 26.1          & 31.3          & 62.0          & 67.2          & 52.1          & 88.2          & 59.8 \\
MBDF-Net \cite{tan2021mbdf}      & LiDAR + RGB & \textbf{81.5}         & 84.7          & 33.0          & 77.3          & 31.2          & 29.0          & 57.7         & 65.6          & 49.9         & 85.5          & 59.5 
\\\bottomrule
ImLiDAR     & LiDAR + RGB & 80.3          & 85.3 & 35.7         & \textbf{79.4} & \textbf{35.4}          & \textbf{38.4}          & 65.9 & 69.9 & \textbf{54.2} & 91.6          & \textbf{63.6} \\\bottomrule
\end{tabular}
\label{compare1}
\end{table*}

\section{Experiments}\label{experiment}
\textbf{Dataset and metric.} We conduct experiments on the KITTI dataset \cite{geiger2012we} and the SUN-RGBD dataset \cite{song2015sun}. KITTI is an outdoor standard benchmark dataset, which consists of 7,481 frames for training and 7,518 frames for testing. 
Following the protocol of \cite{qi2018frustum,shi2019pointrcnn}, we split the 7,481 frames into 3,712 frames for training and 3,769 frames for validation. Three levels of difficulty are defined in the benchmark according to size, occlusion, and truncation, i.e., Easy, Moderate, and Hard. 
Besides, our results are reported for the car, pedestrian and cyclist categories and the IoU thresholds are set to 0.7, 0.5, and 0.5, respectively. 
SUN-RGBD is an indoor benchmark dataset, which includes 10,335 images with 700 annotated object categories, including 5,285 images for training and 5,050 images for testing. The IoU thresholds for all ten categories are set to 0.25. The common Average Precision (AP) is used as our evaluation metric following the official evaluation protocol of the KITTI dataset and the SUN-RGBD dataset. Especially, the 40 recall positions-based metric $AP|R40$ has been utilized by the KITTI dataset instead of $AP|R11$ as before.

\textbf{Implementation details.}  Each LiDAR point cloud is cropped to the range of [-40, 40], [-1, 3], [0, 70.4] meters along (X, Y, Z) axes in the camera coordinate, respectively. The orientation of $\theta$ is set to the range of [$-\pi$, $\pi$]. Similar to PointRCNN \cite{shi2019pointrcnn}, we subsample 16,384 points from each 3D point cloud scene as the inputs of the point stream. Then four set abstraction layers with multi-scale grouping are used to subsample the aforementioned input points into groups with the sizes of 4096, 1024, 256, 64 respectively, and four feature propagation layers are employed to obtain the per-point feature vectors. The image stream takes camera images of the size of 1280 × 384 as input.

\textbf{Training details.} ImLiDAR is trained by SGD with an initial learning rate being 0.002, the momentum being 0.9, and the weight decay being 0.001 respectively. We train the model for around 50 epochs on an Nvidia GeForce RTX 3090 GPU with a batch size of 2 in an end-to-end manner. The balancing weight $\lambda$ in the loss function is set to 1.
\begin{figure*}[!t] \centering
	\includegraphics[width=1.0\linewidth]{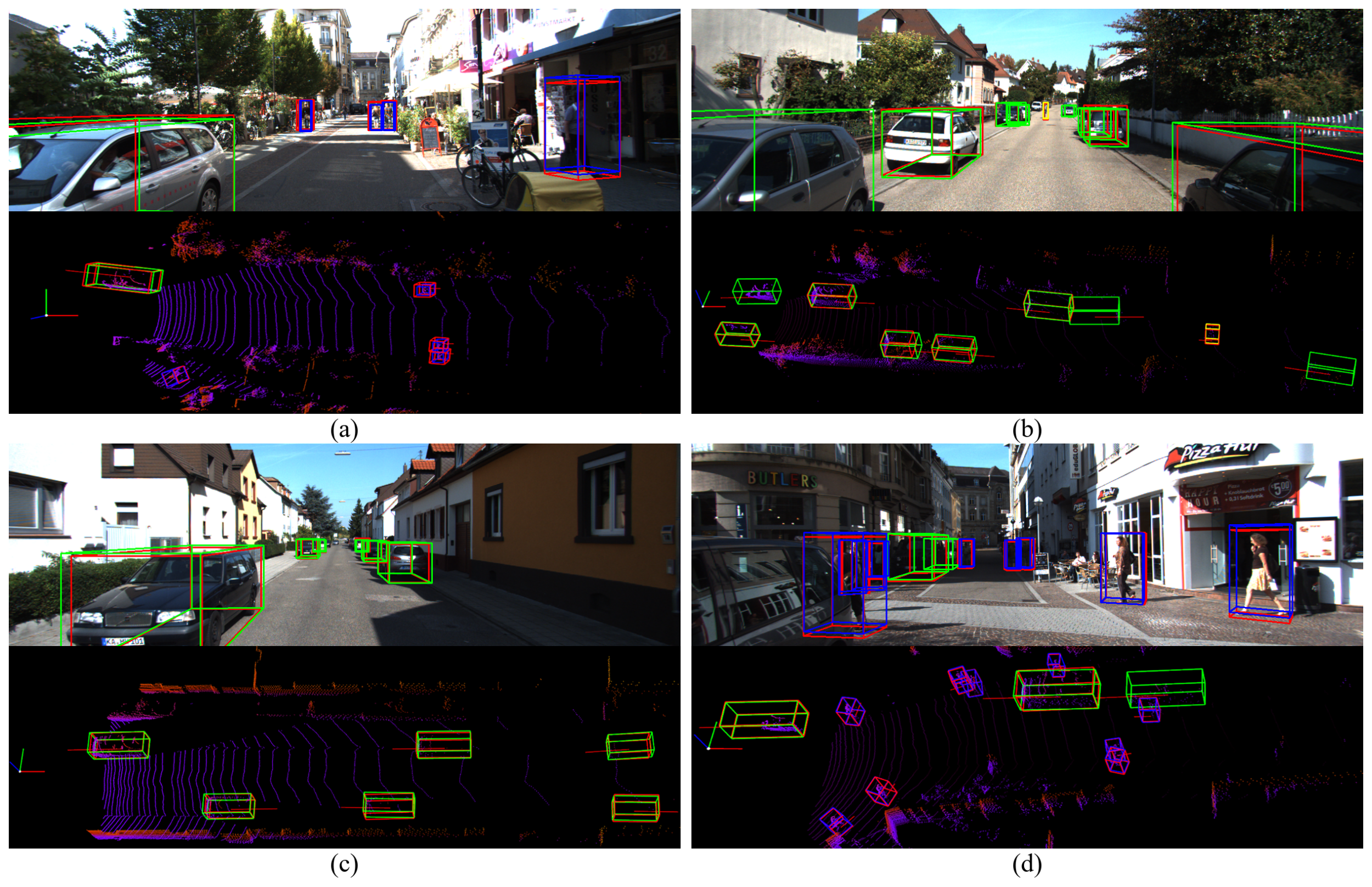}
	\caption{Qualitative results on the val set of the KITTI dataset. We show our detected results of four different scenes in (a)–(d), in which car, pedestrian, and cyclist are shown in \textcolor{green}{green}, \textcolor{blue}{blue}, and \textcolor{yellow}{yellow}, respectively. Note that all ground truth bounding boxes are shown in \textcolor{red}{red}.
	}
	\label{fig:Car}
\end{figure*}
\subsection{Comparison with the State-of-the-Arts}
We evaluate our ImLiDAR with state-of-the-art 3D object detection methods on the KITTI test set (see Table \ref{compare}) and the SUN-RGBD test set (see Table \ref{compare1}). 
As shown in Table \ref{compare}, the point cloud-based methods outperform most of the cross-sensor methods, indicating that fusing the representations of camera images and LiDAR point clouds remains a challenging task. 
While ImLiDAR achieves remarkable results over the state-of-the-art methods on all three categories. For the car category, we improve the baseline PointRCNN \cite{shi2019pointrcnn} by 6.20\% on the mAP metric, and ImLiDAR outperforms all point cloud-based methods on the mAP metric. 
Further, ImLiDAR surpasses all the cross-sensor methods
by a large margin. For example, ImLiDAR outperforms EPNet \cite{huang2020epnet} by 1.17\%, 3.95\%, and 3.08\% on the easy, moderate, and hard metrics respectively, which demonstrates the superiority of the CDMP module. 
It is noteworthy that ImLiDAR also ranks the first on the pedestrian and cyclist categories, although most of existing approaches do not provide evaluations on these two categories. 
The small or partial instances in these categories require more context information with large receptive fields, which can be fully gained by the CDMP module. 
We also provide qualitative detection results on the KITTI validation dataset in Fig. \ref{fig:Car}, from which we can see that ImLiDAR can detect more hard examples, even occluded and distant instances in crowded scenes.

% our method achieves remarkable results. For point cloud based methods, we improve the baseline PointRCNN \cite{shi2019pointrcnn} by 5.51\% on the mAP metric and outperform all methods on the mAP metric. For LiDAR+RGB methods, our ImLiDAR achieves better performance over the state-of-the-art methods for all difficulty levels. Especially, the proposed ImLiDAR outperforms EPNet \cite{huang2020epnet} by 0.67\%, 3.01\%, and 2.58\% on the easy, moderate, and hard metrics respectively, which demonstrate the superiority of the CDMP module. Besides, we also provide qualitative detection results on the KITTI validation split in Fig. \ref{fig:Car}.
To verify the effectiveness of all methods in the indoor scenes, we compare ImLiDAR with its competitors on the SUN-RGBD dataset in Table \ref{compare1}. 
It is noteworthy that our ImLiDAR still outperforms its competitors.
Especially, PointFusion \cite{xu2018pointfusion} and F-PointNet \cite{qi2018frustum} generate 2D bounding boxes from camera images using 2D detectors and output the 3D boxes in a cascading manner.
While ImLiDAR does not add explicit supervision information (e.g., annotations of 2D detection boxes), and outperforms them by 19.5\% and 9.6\% mAP, respectively.
EPNet \cite{huang2020epnet} is a two-branch detector, which directly fuses point clouds and camera images, and the following work \cite{tan2021mbdf} designs a multi-branch fusion manner. 
Our ImLiDAR outperforms EPNet \cite{huang2020epnet} and MBDF-Net \cite{tan2021mbdf} by 3.8\% and 4.1\% in terms of 3D mAP. Such a large improvement verifies the superiority of our CDMP module over the other fusion schemes.

\subsection{Ablation Study}
We conduct extensive experiments on the KITTI validation dataset to evaluate the effectiveness of our CDMP module and the set-based detector.
\begin{table}[]
\small
\centering
\caption{Analysis of different fusion manners on the KITTI val set (car). 
Note that SC and AD represent the simple concatenation and addition of point features and point-wise image features, respectively. 
CDMP and CDMP* represent the CDMP ($1\times1$) modules in the set abstraction layers and the CDMP ($1\times4$) module in the last feature propagation layer. LI and LI* denote LI-Fusion modules \cite{huang2020epnet} in similar architectures. It should be noted that no image stream is employed for the baseline (the first row), and all models adopt the NMS procedure to keep more accurate bounding boxes.}
\setlength{\tabcolsep}{0.2mm}{
\begin{tabular}{ccccccccccc}
\toprule
\multicolumn{6}{c}{Fusion}      & \multicolumn{5}{c}{3D Detection}         \\ 
SC & AD & LI & LI* & CDMP & CDMP* & Easy  & Moderate & Hard  & 3D mAP & Gain \\ \midrule
\XSolid&\XSolid    &\XSolid    &\XSolid    &\XSolid      &\XSolid      & 86.24 & 77.36 & 75.88 & 79.82  & -    \\ 
\Checkmark  &\XSolid     &\XSolid&\XSolid&\XSolid  &\XSolid  & 85.68 & 76.78    & 75.26 & 79.24  &{\color{green}$\downarrow 0.58$}\\ 
\XSolid   &\Checkmark     &\XSolid&\XSolid&\XSolid  &\XSolid  & 84.10 & 73.59    & 71.97 & 76.55  &{\color{green}$\downarrow 3.26$}\\ 
\XSolid   &\XSolid&\Checkmark     &\XSolid&\XSolid  &\XSolid  & 87.17 & 78.31    & 76.10 & 80.52 &{\color{red}$\uparrow 0.70$} \\ 
\XSolid   &\XSolid&\XSolid&\Checkmark     &\XSolid  &\XSolid  & 86.45 & 77.94    & 76.39 & 80.26  &{\color{red}$\uparrow 0.44$}\\ 
\XSolid   &\XSolid&\Checkmark     &\Checkmark     &\XSolid  &\XSolid  & 89.26 & 78.88    & 76.82 & 81.65  &{\color{red}$\uparrow  1.83$}\\ 
\XSolid   &\XSolid&\XSolid&\XSolid&\Checkmark       &\XSolid  & 90.42 & 81.84    & 79.38 & 83.88  &{\color{red}$\uparrow 4.06$} \\ 
\XSolid   &\XSolid&\XSolid&\XSolid&\XSolid  &\Checkmark       & 90.32 & 80.93    & 78.72 & 83.32  &{\color{red}$\uparrow 3.50$} \\ 
\XSolid   &\XSolid&\XSolid&\XSolid&\Checkmark       &\Checkmark & 91.89 & 83.38  & 81.62 & 85.63  &{\color{red}$\uparrow 5.81$} \\  \bottomrule
\end{tabular}
}
\label{ab1}
\end{table}

\textbf{Effectiveness of the CDMP module.} 
We conduct some ablation experiments on the CDMP module. 
For fair comparisons, all the models adopt the same NMS procedure for filtering out low-quality proposals. 
Table \ref{ab1} shows the results of different fusion modules. 
It is found that: (1) Simple concatenation (SC) and addition (AD) yield the decrease of 3D mAP 0.58\% and 3.26\% over the baseline, which indicates that such simple fusion manners cannot obtain more accurate 3D detection results than only using LiDAR data, and even worse. 
(2) The combination of LI and LI*, along with the combination of CDMP and CDMP*, performs better than single-scale fusion modules, which verifies the effectiveness of cross-sensor fusion in multiple scales and stages. 
(3) The combination of CDMP and CDMP* modules yields the most significant improvement of 5.81\% in terms of 3D mAP, demonstrating that our CDMP modules actually provide a more effective way to fuse the multi-scale image and point features, thus leading to more quality 3D object detection results than the LI-Fusion modules \cite{huang2020epnet}.

% (1) SC and AD yields decreasements of 3D mAP 0.58\% and 3.26\% over the baseline, which indicates that such simple fusion manners cannot obtain more accurate 3D detection results than only using LiDAR data, and even worse. (2) the combination of LI and LI*, along with the combination of CDMP and CDMP* performer better than only utilizing one typical fusion module, which verifies the effectiveness of fusion in multiple scales. (3) Our CDMP modules yield the most significant improvement of 4.89\% in terms of 3D mAP, demonstrating that our CDMP actually provide a more effective way to fuse the multi-scale point information with the image context, thus leading to accurate 3D detection, compared with LI-Fusion modules \cite{huang2020epnet}.
\begin{figure}[!t] \centering
	\includegraphics[width=1\linewidth]{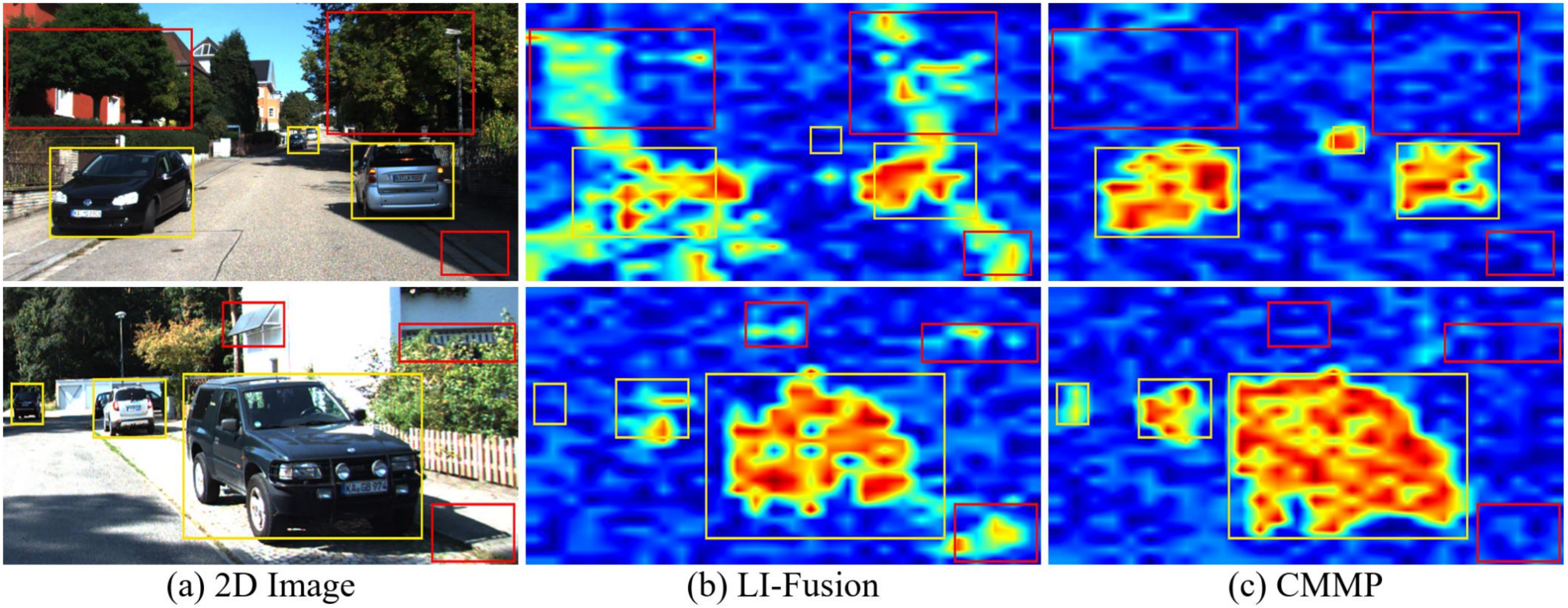}
	\caption{Visualization of the semantic image feature. Foreground objects are highlighted with yellow rectangle boxes. The red rectangle box marks the interfering image features.}
	\label{fig:semantic}
\end{figure}

\textbf{Visualization of semantic image features.} How to effectively deliver useful semantic information to enrich the point features is critical for cross-sensor 3D detection. 
We compare the image features learned from the LI-Fusion module \cite{huang2020epnet} and our CDMP module as shown in Fig. \ref{fig:semantic}. 
Although the LI-Fusion modules attempt to suppress the bad information by assigning coarse learnable weight matrices, they achieve a limited effect and still retain much interfering image information.
Besides, the limited receptive fields will cause the loss of important semantic information. 
Comparatively, our CDMP modules can effectively gather long-range key context information and suppress harmful semantic information, thus leading to more accurate detection results.

\begin{table}[]
\small
\centering
\caption{Analysis of the set-based detector on the KITTI val set.}
\setlength{\tabcolsep}{0.4mm}{
\begin{tabular}{ccccccccc}
\toprule
\multicolumn{4}{c}{Model}   & \multicolumn{5}{c}{3D Detection (Car)}                        \\
IoU & CE  & Set-based & NMS & Easy  & Moderate                     & Hard  & 3D mAP & Gain  \\\midrule
\XSolid & \XSolid & \XSolid       & \Checkmark & 91.89 & 83.38  & 81.62 & 85.63  & -     \\
\Checkmark & \XSolid & \XSolid       & \Checkmark & 91.38 & 83.47  & 81.89 & 85.58  & {\color{green}$\downarrow0.05$} \\
\Checkmark & \XSolid & \XSolid       & \XSolid & 87.57 & 78.59 & 76.33 & 80.83  & {\color{green}$\downarrow4.80 $} \\
\XSolid & \Checkmark & \XSolid       & \Checkmark & 92.21 & 83.23  & 81.79 & 85.74  & {\color{red}$\uparrow0.11$} \\
\XSolid & \XSolid  & \XSolid       & \Checkmark & 88.12 & 79.70 & 78.01 & 81.94  & {\color{green}$\downarrow3.68 $}  \\
\XSolid & \XSolid & \Checkmark       & \XSolid & 92.61 & 85.52  & 83.25 & 87.13  & {\color{red}$\uparrow1.50$} \\
\XSolid & \XSolid & \Checkmark       & \Checkmark & 92.66 & 85.51  & 83.24 & 87.14  & {\color{red}$\uparrow1.51$} \\\bottomrule
\multicolumn{4}{c}{PointRCNN \cite{shi2019pointrcnn}} & 89.19 & 78.85 & 77.91 & 81.98  & - \\
\multicolumn{4}{c}{PointRCNN + Set-based} & 91.09 & 80.31 & 78.67 & 83.35  & {\color{red}$\uparrow1.37$} \\\bottomrule
\multicolumn{4}{c}{EPNet \cite{huang2020epnet}} & 92.17 & 82.68 & 80.10 & 84.98  & - \\
\multicolumn{4}{c}{EPNet + Set-based} & 92.49 & 84.06& 81.19 & 85.91  & {\color{red}$\uparrow0.93$} \\\bottomrule
\end{tabular}
}
% $vs.$ NMS procudure, IoU \cite{yu2016unitbox} and CE \cite{huang2020epnet} loss functions. The balancing weights in IoU and CE loss functions both are set to 5.}
\label{ab2}
\end{table}

\begin{figure}[!t] \centering
	\includegraphics[width=1\linewidth]{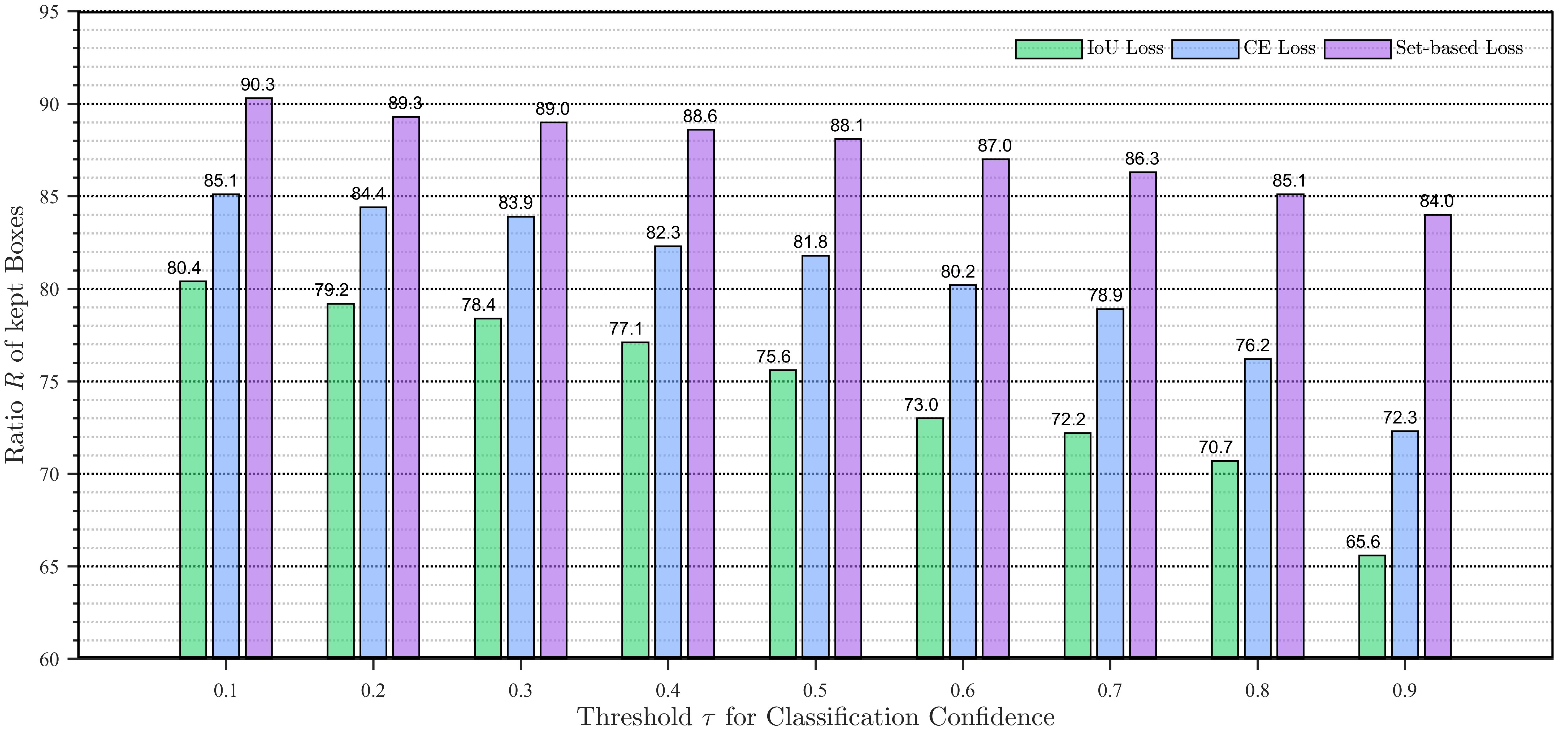}
	\caption{Illustration of the ratio of keeping positive boxes with different classification confidence thresholds.}
	\label{fig:ratio}
\end{figure}

\begin{figure}[!t] \centering
	\includegraphics[width=1\linewidth]{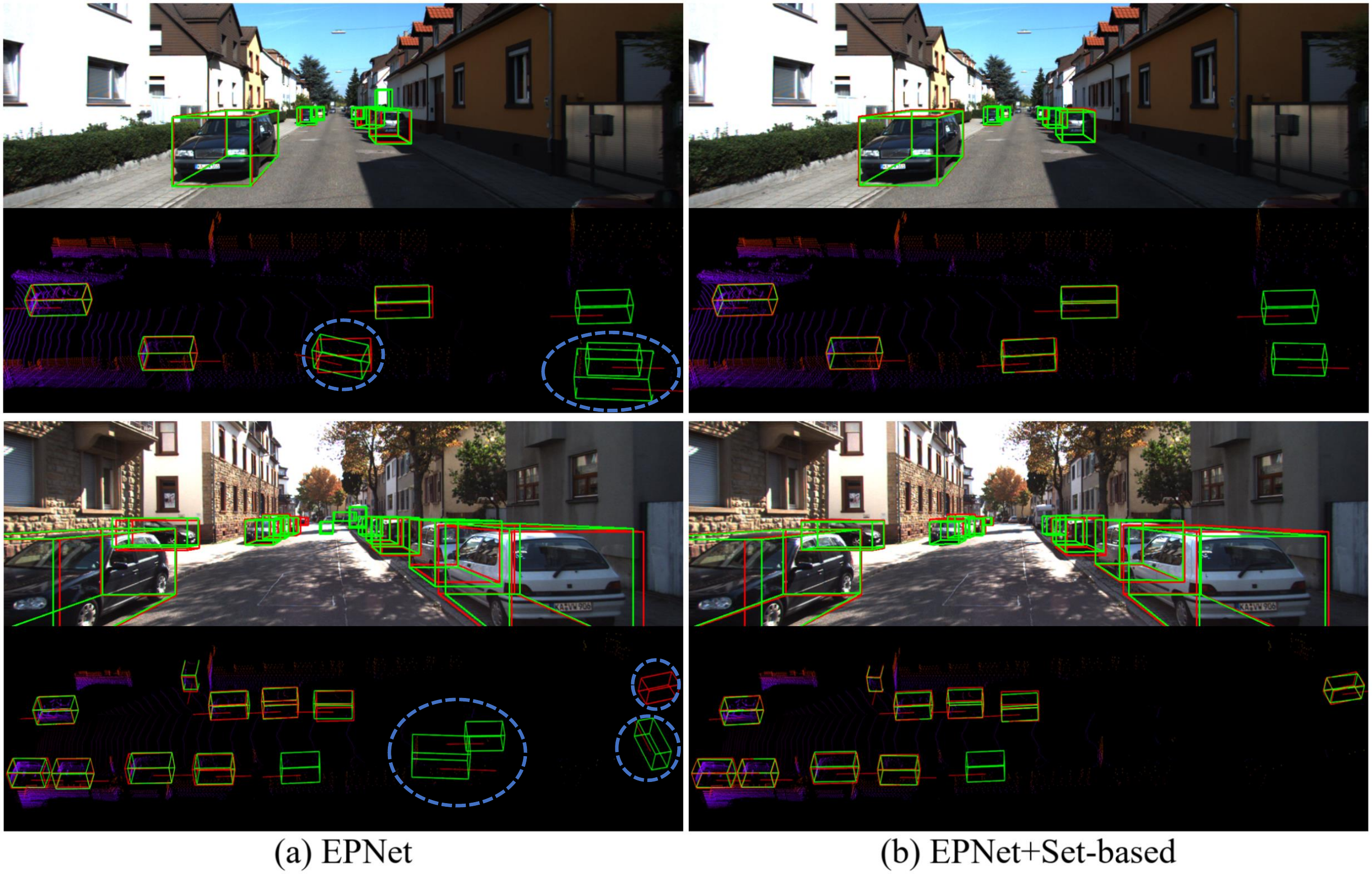}
	\caption{Visualization results by EPNet \cite{huang2020epnet} and the combination of EPNet \cite{huang2020epnet} and our set-based detector. It is noteworthy that our set-based detector can filter out false positives, avoid missing distant objects, and even improve the predicted results.}
	\label{fig:qa_ab}
\end{figure}

\begin{table}[!t]
\small
\centering
\caption{Analysis of the set-based detector on the SUN-RGBD test set.}
\begin{tabular}{cccc}
\toprule
Model     & chair & desk                        & table \\\midrule
Baseline  & 73.3  & 27.1                        & 50.0  \\
IoU       & 74.8  & 28.2                        & 51.4  \\
CE        & 75.7  & 28.5                        & 51.8  \\
Set-based & 79.4  & 35.4                        & 54.2  \\\bottomrule
\end{tabular}
\label{ab4}
\end{table}

% \begin{table}[]
% \small
% \centering
% \begin{tabular}{|c|c|c|c|}
% \hline
% Model     & \multicolumn{3}{|c|}{3D Detection (SUN-RGBD)} \\\hline
%           & chair & desk                        & table \\\hline
% Baseline  & 68.3  & 20.1                        & 47.0  \\\hline
% IoU       & 68.8  & 19.2                        & 47.4  \\\hline
% CE        & 75.7  & 25.5                        & 52.8  \\\hline
% Set-based & 77.6  & 27.3                        & 54.2  \\\hline
% \end{tabular}
% \end{table}

\textbf{Effectiveness of the set-based detector.} In Table \ref{ab2}, the set-based detector is evaluated with CE and IoU loss functions. The balancing weights in IoU and CE loss functions are also set to 5. 
In the car category, the set-based detector yields a significant improvement of 1.50\% mAP over the baseline, which indicates the superiority of our set-based detector in improving the 3D detection performance. 
Besides, we also compare the set-based detector, CE loss function, and IoU loss function with and without NMS. 
Without NMS, the set-based detector drops in performance by only 0.01\% while CE and IoU drop by 3.79\% and 4.75\% respectively. It shows that our set-based detector still works well even without NMS. 
Moreover, we combine PointRCNN \cite{shi2019pointrcnn} and EPNet \cite{huang2020epnet} with the set-based detector. It indicates that the set-based detector is beneficial for generating more high-quality proposals even without the NMS post-processing. As shown in Fig. \ref{fig:qa_ab}, our set-based detector can filter out false positives, avoid missing distant objects, and even improve the predicted results.

Following the protocol of EPNet \cite{huang2020epnet}, we adopt the ratio of $\mathcal{R}$ to figure out how the consistency between these two confidences is improved, which is formulated as:
\begin{equation}
\mathcal{R} = \frac{\mathcal{N}(b|b\in \mathcal{B}\, and\, c_{b}>v)}{\mathcal{N}(\mathcal{B})}
\end{equation}
where $\mathcal{B}$ denotes the set of positive candidate boxes, which are filtered by a predefined IoU threshold $\tau$. And following \cite{huang2020epnet}, we set $\tau$ to 0.7,
$c_{b}$ represents the classification confidence of the positive candidate box $b$, and $v$ is another threshold to filter positive candidate boxes with smaller classification confidence. $\mathcal{N}(.)$ calculates the number of boxes. 
In all different settings of classification confidence threshold $v$, all models generate 64 boxes without the NMS procedure employed. 
These boxes are used to get the positive candidate boxes by calculating the overlaps with the ground truth boxes. As shown in Fig. \ref{fig:ratio}, the model with the set-based detector demonstrates better consistency than that trained with IoU loss and CE loss functions. 
Further, we evaluate different models on the SUN-RGBD test set. Especially, we select three categories ``chair'', ``desk'' and ``table'' in crowded scenes. As shown in Table \ref{ab4}, the proposed set-based detector still outperforms other models in these categories. Such large gains demonstrate that our set-based detector performs better post-processing, keeping more highly-overlapped true positives, especially in the crowded scene.
\begin{figure}[!t] \centering
	\includegraphics[width=1.0\linewidth]{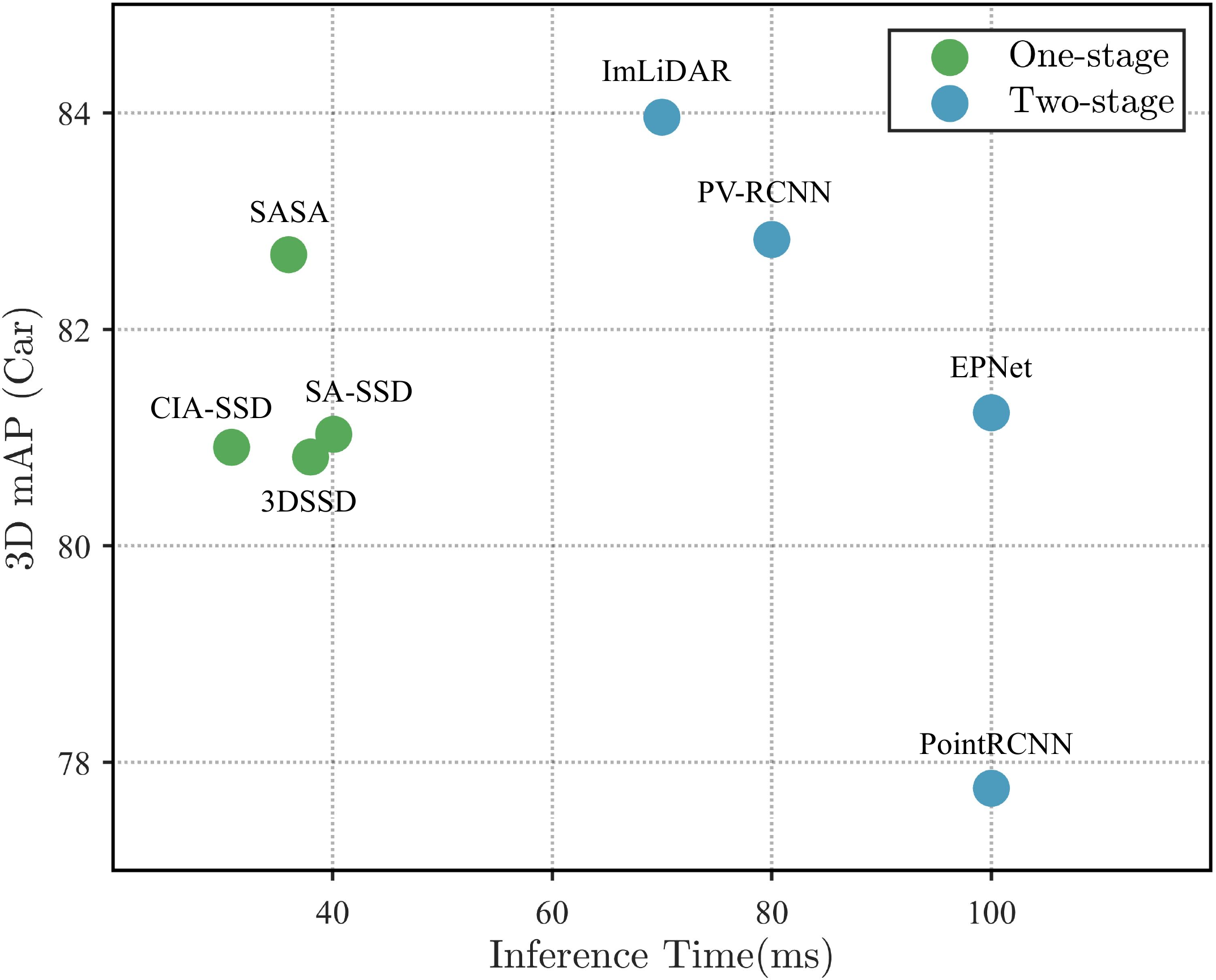}
	\caption{Comparisons on inference time (ms) on the KITTI dataset. Our method reaches top performance among both one-stage and two-stage detectors.
	}
	\label{fig:time}
\end{figure}
\subsection{Limitations}
Fig. \ref{fig:time} shows the inference time among several compared methods. Our two-stage framework takes around 70ms to process a single point cloud sample from the KITTI dataset on an RTX 3090 GPU. Our model needs more time than one-stage frameworks, but it surpasses all one-stage detectors by a great margin. Meanwhile, our model is significantly faster
than all two-stage detectors. This is because our model bypasses the consuming calculation of the NMS post-processing procedure. Besides, our model still obtains much higher mAP compared with two-stage detectors.

\section{Conclusion}
In this paper, we present ImLiDAR, a novel 3D object detection paradigm, which progressively fuses camera images and LiDAR point clouds in multiple scales for quality 3D object detection.
Two core designs exist in ImLiDAR. First, we present a cross-sensor dynamic message propagation module to combine the best of image and point features. Second, we design a set-based detector to select high-quality bounding boxes with both high classification and localization. It can be easily implemented in any detection network.
Moreover, ImLiDAR does not require additional image annotations, the complex BEV data, and the commonly used NMS post-processing step.
% We also design a set-based detector to select high-quality bounding boxes with both high classification and localization. It also can be easily implemented in any detection network without additional NMS post-processing.
% semantic image features into the point features
% We have proposed a novel cross-sensor dynamic message propagation network (ImLiDAR) for 3D object detection. ImLiDAR consists of a two-stream RPN, a set-based detector, and a refinement network. The two-stream RPN can effectively fuse semantic image features into the point features by introducing the proposed CDMP modules. Besides, we design a set-based detector to resolve the problems resulting from the NMS procedure. Concretely, the high-quality proposals with high classification and localization confidence are kept by the proposed set-based detector without NMS post-processing. Finally, these high-quality proposals are fed into the refinement network for bounding box refining. 
Extensive experiments on KITTI and SUN-RGBD datasets verify the superiority of ImLiDAR. 

% In the future, we would like to investigate a more effective cross-sensor fusion paradigm to fuse more different types of sensors.

\bibliographystyle{IEEEtran}
\bibliography{acmart}

\begin{thebibliography}{10}
\providecommand{\url}[1]{#1}
\csname url@samestyle\endcsname
\providecommand{\newblock}{\relax}
\providecommand{\bibinfo}[2]{#2}
\providecommand{\BIBentrySTDinterwordspacing}{\spaceskip=0pt\relax}
\providecommand{\BIBentryALTinterwordstretchfactor}{4}
\providecommand{\BIBentryALTinterwordspacing}{\spaceskip=\fontdimen2\font plus
\BIBentryALTinterwordstretchfactor\fontdimen3\font minus
  \fontdimen4\font\relax}
\providecommand{\BIBforeignlanguage}[2]{{%
\expandafter\ifx\csname l@#1\endcsname\relax
\typeout{** WARNING: IEEEtran.bst: No hyphenation pattern has been}%
\typeout{** loaded for the language `#1'. Using the pattern for}%
\typeout{** the default language instead.}%
\else
\language=\csname l@#1\endcsname
\fi
#2}}
\providecommand{\BIBdecl}{\relax}
\BIBdecl

\bibitem{chen2016monocular}
X.~Chen, K.~Kundu, Z.~Zhang, H.~Ma, S.~Fidler, and R.~Urtasun, ``Monocular 3d
  object detection for autonomous driving,'' in \emph{Proceedings of the IEEE
  conference on computer vision and pattern recognition}, 2016, pp. 2147--2156.

\bibitem{wang2021depth}
L.~Wang, L.~Du, X.~Ye, Y.~Fu, G.~Guo, X.~Xue, J.~Feng, and L.~Zhang,
  ``Depth-conditioned dynamic message propagation for monocular 3d object
  detection,'' in \emph{Proceedings of the IEEE/CVF Conference on Computer
  Vision and Pattern Recognition}, 2021, pp. 454--463.

\bibitem{chabot2017deep}
F.~Chabot, M.~Chaouch, J.~Rabarisoa, C.~Teuliere, and T.~Chateau, ``Deep manta:
  A coarse-to-fine many-task network for joint 2d and 3d vehicle analysis from
  monocular image,'' in \emph{Proceedings of the IEEE conference on computer
  vision and pattern recognition}, 2017, pp. 2040--2049.

\bibitem{chen20173d}
X.~Chen, K.~Kundu, Y.~Zhu, H.~Ma, S.~Fidler, and R.~Urtasun, ``3d object
  proposals using stereo imagery for accurate object class detection,''
  \emph{IEEE transactions on pattern analysis and machine intelligence},
  vol.~40, no.~5, pp. 1259--1272, 2017.

\bibitem{li2019stereo}
P.~Li, X.~Chen, and S.~Shen, ``Stereo r-cnn based 3d object detection for
  autonomous driving,'' in \emph{Proceedings of the IEEE/CVF Conference on
  Computer Vision and Pattern Recognition}, 2019, pp. 7644--7652.

\bibitem{chen20153d}
X.~Chen, K.~Kundu, Y.~Zhu, A.~G. Berneshawi, H.~Ma, S.~Fidler, and R.~Urtasun,
  ``3d object proposals for accurate object class detection,'' \emph{Advances
  in neural information processing systems}, vol.~28, 2015.

\bibitem{luo2018fast}
W.~Luo, B.~Yang, and R.~Urtasun, ``Fast and furious: Real time end-to-end 3d
  detection, tracking and motion forecasting with a single convolutional net,''
  in \emph{Proceedings of the IEEE conference on Computer Vision and Pattern
  Recognition}, 2018, pp. 3569--3577.

\bibitem{yang2018pixor}
B.~Yang, W.~Luo, and R.~Urtasun, ``Pixor: Real-time 3d object detection from
  point clouds,'' in \emph{Proceedings of the IEEE conference on Computer
  Vision and Pattern Recognition}, 2018, pp. 7652--7660.

\bibitem{zhou2018voxelnet}
Y.~Zhou and O.~Tuzel, ``Voxelnet: End-to-end learning for point cloud based 3d
  object detection,'' in \emph{Proceedings of the IEEE conference on computer
  vision and pattern recognition}, 2018, pp. 4490--4499.

\bibitem{qi2018frustum}
C.~R. Qi, W.~Liu, C.~Wu, H.~Su, and L.~J. Guibas, ``Frustum pointnets for 3d
  object detection from rgb-d data,'' in \emph{Proceedings of the IEEE
  conference on computer vision and pattern recognition}, 2018, pp. 918--927.

\bibitem{liang2019multi}
M.~Liang, B.~Yang, Y.~Chen, R.~Hu, and R.~Urtasun, ``Multi-task multi-sensor
  fusion for 3d object detection,'' in \emph{Proceedings of the IEEE/CVF
  Conference on Computer Vision and Pattern Recognition}, 2019, pp. 7345--7353.

\bibitem{huang2020epnet}
T.~Huang, Z.~Liu, X.~Chen, and X.~Bai, ``Epnet: Enhancing point features with
  image semantics for 3d object detection,'' in \emph{European Conference on
  Computer Vision}.\hskip 1em plus 0.5em minus 0.4em\relax Springer, 2020, pp.
  35--52.

\bibitem{zhao20193d}
X.~Zhao, Z.~Liu, R.~Hu, and K.~Huang, ``3d object detection using scale
  invariant and feature reweighting networks,'' in \emph{Proceedings of the
  AAAI Conference on Artificial Intelligence}, vol.~33, no.~01, 2019, pp.
  9267--9274.

\bibitem{xu2018pointfusion}
D.~Xu, D.~Anguelov, and A.~Jain, ``Pointfusion: Deep sensor fusion for 3d
  bounding box estimation,'' in \emph{Proceedings of the IEEE conference on
  computer vision and pattern recognition}, 2018, pp. 244--253.

\bibitem{liang2018deep}
M.~Liang, B.~Yang, S.~Wang, and R.~Urtasun, ``Deep continuous fusion for
  multi-sensor 3d object detection,'' in \emph{Proceedings of the European
  conference on computer vision (ECCV)}, 2018, pp. 641--656.

\bibitem{chen2017multi}
X.~Chen, H.~Ma, J.~Wan, B.~Li, and T.~Xia, ``Multi-view 3d object detection
  network for autonomous driving,'' in \emph{Proceedings of the IEEE conference
  on Computer Vision and Pattern Recognition}, 2017, pp. 1907--1915.

\bibitem{ku2018joint}
J.~Ku, M.~Mozifian, J.~Lee, A.~Harakeh, and S.~L. Waslander, ``Joint 3d
  proposal generation and object detection from view aggregation,'' in
  \emph{2018 IEEE/RSJ International Conference on Intelligent Robots and
  Systems (IROS)}.\hskip 1em plus 0.5em minus 0.4em\relax IEEE, 2018, pp. 1--8.

\bibitem{tan2021mbdf}
X.~Tan, X.~Chen, G.~Zhang, J.~Ding, and X.~Lan, ``Mbdf-net: Multi-branch deep
  fusion network for 3d object detection,'' in \emph{Proceedings of the 1st
  International Workshop on Multimedia Computing for Urban Data}, 2021, pp.
  9--17.

\bibitem{huang2021joint}
K.~Huang and Q.~Hao, ``Joint multi-object detection and tracking with
  camera-lidar fusion for autonomous driving,'' in \emph{2021 IEEE/RSJ
  International Conference on Intelligent Robots and Systems (IROS)}.\hskip 1em
  plus 0.5em minus 0.4em\relax IEEE, 2021, pp. 6983--6989.

\bibitem{piergiovanni20214d}
A.~Piergiovanni, V.~Casser, M.~S. Ryoo, and A.~Angelova, ``4d-net for learned
  multi-modal alignment,'' in \emph{Proceedings of the IEEE/CVF International
  Conference on Computer Vision}, 2021, pp. 15\,435--15\,445.

\bibitem{liu2021epnet++}
Z.~Liu, B.~Li, X.~Chen, X.~Wang, X.~Bai \emph{et~al.}, ``Epnet++: Cascade
  bi-directional fusion for multi-modal 3d object detection,'' \emph{arXiv
  preprint arXiv:2112.11088}, 2021.

\bibitem{yoo20203d}
J.~H. Yoo, Y.~Kim, J.~Kim, and J.~W. Choi, ``3d-cvf: Generating joint camera
  and lidar features using cross-view spatial feature fusion for 3d object
  detection,'' in \emph{European Conference on Computer Vision}.\hskip 1em plus
  0.5em minus 0.4em\relax Springer, 2020, pp. 720--736.

\bibitem{ku2019monocular}
J.~Ku, A.~D. Pon, and S.~L. Waslander, ``Monocular 3d object detection
  leveraging accurate proposals and shape reconstruction,'' in
  \emph{Proceedings of the IEEE/CVF conference on computer vision and pattern
  recognition}, 2019, pp. 11\,867--11\,876.

\bibitem{liu2019deep}
L.~Liu, J.~Lu, C.~Xu, Q.~Tian, and J.~Zhou, ``Deep fitting degree scoring
  network for monocular 3d object detection,'' in \emph{Proceedings of the
  IEEE/CVF Conference on Computer Vision and Pattern Recognition}, 2019, pp.
  1057--1066.

\bibitem{reading2021categorical}
C.~Reading, A.~Harakeh, J.~Chae, and S.~L. Waslander, ``Categorical depth
  distribution network for monocular 3d object detection,'' in
  \emph{Proceedings of the IEEE/CVF Conference on Computer Vision and Pattern
  Recognition}, 2021, pp. 8555--8564.

\bibitem{wang2019pseudo}
Y.~Wang, W.-L. Chao, D.~Garg, B.~Hariharan, M.~Campbell, and K.~Q. Weinberger,
  ``Pseudo-lidar from visual depth estimation: Bridging the gap in 3d object
  detection for autonomous driving,'' in \emph{Proceedings of the IEEE/CVF
  Conference on Computer Vision and Pattern Recognition}, 2019, pp. 8445--8453.

\bibitem{li2019gs3d}
B.~Li, W.~Ouyang, L.~Sheng, X.~Zeng, and X.~Wang, ``Gs3d: An efficient 3d
  object detection framework for autonomous driving,'' in \emph{Proceedings of
  the IEEE/CVF Conference on Computer Vision and Pattern Recognition}, 2019,
  pp. 1019--1028.

\bibitem{2018Multi}
B.~Xu and Z.~Chen, ``Multi-level fusion based 3d object detection from
  monocular images,'' in \emph{2018 IEEE/CVF Conference on Computer Vision and
  Pattern Recognition (CVPR)}, 2018.

\bibitem{ding2020learning}
M.~Ding, Y.~Huo, H.~Yi, Z.~Wang, J.~Shi, Z.~Lu, and P.~Luo, ``Learning
  depth-guided convolutions for monocular 3d object detection,'' in
  \emph{Proceedings of the IEEE/CVF Conference on computer vision and pattern
  recognition workshops}, 2020, pp. 1000--1001.

\bibitem{chen2020dsgn}
Y.~Chen, S.~Liu, X.~Shen, and J.~Jia, ``Dsgn: Deep stereo geometry network for
  3d object detection,'' in \emph{Proceedings of the IEEE/CVF conference on
  computer vision and pattern recognition}, 2020, pp. 12\,536--12\,545.

\bibitem{lang2019pointpillars}
A.~H. Lang, S.~Vora, H.~Caesar, L.~Zhou, J.~Yang, and O.~Beijbom,
  ``Pointpillars: Fast encoders for object detection from point clouds,'' in
  \emph{Proceedings of the IEEE/CVF Conference on Computer Vision and Pattern
  Recognition}, 2019, pp. 12\,697--12\,705.

\bibitem{deng2020voxel}
J.~Deng, S.~Shi, P.~Li, W.~Zhou, Y.~Zhang, and H.~Li, ``Voxel r-cnn: Towards
  high performance voxel-based 3d object detection,'' \emph{arXiv preprint
  arXiv:2012.15712}, vol.~1, no.~2, p.~4, 2020.

\bibitem{shi2019pointrcnn}
S.~Shi, X.~Wang, and H.~Li, ``Pointrcnn: 3d object proposal generation and
  detection from point cloud,'' in \emph{Proceedings of the IEEE/CVF conference
  on computer vision and pattern recognition}, 2019, pp. 770--779.

\bibitem{qi2019deep}
C.~R. Qi, O.~Litany, K.~He, and L.~J. Guibas, ``Deep hough voting for 3d object
  detection in point clouds,'' in \emph{proceedings of the IEEE/CVF
  International Conference on Computer Vision}, 2019, pp. 9277--9286.

\bibitem{zhang2020pc}
Y.~Zhang, D.~Huang, and Y.~Wang, ``Pc-rgnn: Point cloud completion and graph
  neural network for 3d object detection,'' \emph{arXiv preprint
  arXiv:2012.10412}, 2020.

\bibitem{pan20213d}
X.~Pan, Z.~Xia, S.~Song, L.~E. Li, and G.~Huang, ``3d object detection with
  pointformer,'' in \emph{Proceedings of the IEEE/CVF Conference on Computer
  Vision and Pattern Recognition}, 2021, pp. 7463--7472.

\bibitem{wu2022casa}
H.~Wu, J.~Deng, C.~Wen, X.~Li, C.~Wang, and J.~Li, ``Casa: A cascade attention
  network for 3-d object detection from lidar point clouds,'' \emph{IEEE
  Transactions on Geoscience and Remote Sensing}, vol.~60, pp. 1--11, 2022.

\bibitem{song2016deep}
S.~Song and J.~Xiao, ``Deep sliding shapes for amodal 3d object detection in
  rgb-d images,'' in \emph{Proceedings of the IEEE conference on computer
  vision and pattern recognition}, 2016, pp. 808--816.

\bibitem{yan2018second}
Y.~Yan, Y.~Mao, and B.~Li, ``Second: Sparsely embedded convolutional
  detection,'' \emph{Sensors}, vol.~18, no.~10, p. 3337, 2018.

\bibitem{yu2022siev}
C.~Yu, J.~Lei, B.~Peng, H.~Shen, and Q.~Huang, ``Siev-net: A
  structure-information enhanced voxel network for 3d object detection from
  lidar point clouds,'' \emph{IEEE Transactions on Geoscience and Remote
  Sensing}, 2022.

\bibitem{qi2017pointnet}
C.~R. Qi, H.~Su, K.~Mo, and L.~J. Guibas, ``Pointnet: Deep learning on point
  sets for 3d classification and segmentation,'' in \emph{Proceedings of the
  IEEE conference on computer vision and pattern recognition}, 2017, pp.
  652--660.

\bibitem{qi2017pointnet++}
C.~R. Qi, L.~Yi, H.~Su, and L.~J. Guibas, ``Pointnet++: Deep hierarchical
  feature learning on point sets in a metric space,'' \emph{Advances in neural
  information processing systems}, vol.~30, 2017.

\bibitem{shi2020pv}
S.~Shi, C.~Guo, L.~Jiang, Z.~Wang, J.~Shi, X.~Wang, and H.~Li, ``Pv-rcnn:
  Point-voxel feature set abstraction for 3d object detection,'' in
  \emph{Proceedings of the IEEE/CVF Conference on Computer Vision and Pattern
  Recognition}, 2020, pp. 10\,529--10\,538.

\bibitem{li2021lidar}
Z.~Li, F.~Wang, and N.~Wang, ``Lidar r-cnn: An efficient and universal 3d
  object detector,'' in \emph{Proceedings of the IEEE/CVF Conference on
  Computer Vision and Pattern Recognition}, 2021, pp. 7546--7555.

\bibitem{xie2020pi}
L.~Xie, C.~Xiang, Z.~Yu, G.~Xu, Z.~Yang, D.~Cai, and X.~He, ``Pi-rcnn: An
  efficient multi-sensor 3d object detector with point-based attentive
  cont-conv fusion module,'' in \emph{Proceedings of the AAAI conference on
  artificial intelligence}, vol.~34, no.~07, 2020, pp. 12\,460--12\,467.

\bibitem{du2018general}
X.~Du, M.~H. Ang, S.~Karaman, and D.~Rus, ``A general pipeline for 3d detection
  of vehicles,'' in \emph{2018 IEEE International Conference on Robotics and
  Automation (ICRA)}.\hskip 1em plus 0.5em minus 0.4em\relax IEEE, 2018, pp.
  3194--3200.

\bibitem{scarselli2008graph}
F.~Scarselli, M.~Gori, A.~C. Tsoi, M.~Hagenbuchner, and G.~Monfardini, ``The
  graph neural network model,'' \emph{IEEE transactions on neural networks},
  vol.~20, no.~1, pp. 61--80, 2008.

\bibitem{landrieu2018large}
L.~Landrieu and M.~Simonovsky, ``Large-scale point cloud semantic segmentation
  with superpoint graphs,'' in \emph{Proceedings of the IEEE conference on
  computer vision and pattern recognition}, 2018, pp. 4558--4567.

\bibitem{qi20173d}
X.~Qi, R.~Liao, J.~Jia, S.~Fidler, and R.~Urtasun, ``3d graph neural networks
  for rgbd semantic segmentation,'' in \emph{Proceedings of the IEEE
  International Conference on Computer Vision}, 2017, pp. 5199--5208.

\bibitem{ma2021fast}
F.~Ma, F.~Zhang, Q.~Yin, D.~Xiang, and Y.~Zhou, ``Fast sar image segmentation
  with deep task-specific superpixel sampling and soft graph convolution,''
  \emph{IEEE Transactions on Geoscience and Remote Sensing}, vol.~60, pp.
  1--16, 2021.

\bibitem{zarzar2019pointrgcn}
J.~Zarzar, S.~Giancola, and B.~Ghanem, ``Pointrgcn: Graph convolution networks
  for 3d vehicles detection refinement,'' \emph{arXiv preprint
  arXiv:1911.12236}, 2019.

\bibitem{Shi_2020_CVPR}
W.~Shi and R.~Rajkumar, ``Point-gnn: Graph neural network for 3d object
  detection in a point cloud,'' in \emph{Proceedings of the IEEE/CVF Conference
  on Computer Vision and Pattern Recognition (CVPR)}, June 2020.

\bibitem{tian2021relation}
S.~Tian, L.~Kang, X.~Xing, J.~Tian, C.~Fan, and Y.~Zhang, ``A
  relation-augmented embedded graph attention network for remote sensing object
  detection,'' \emph{IEEE Transactions on Geoscience and Remote Sensing},
  vol.~60, pp. 1--18, 2021.

\bibitem{zhang2020dynamic}
L.~Zhang, D.~Xu, A.~Arnab, and P.~H. Torr, ``Dynamic graph message passing
  networks,'' in \emph{Proceedings of the IEEE/CVF Conference on Computer
  Vision and Pattern Recognition}, 2020, pp. 3726--3735.

\bibitem{dai2017deformable}
J.~Dai, H.~Qi, Y.~Xiong, Y.~Li, G.~Zhang, H.~Hu, and Y.~Wei, ``Deformable
  convolutional networks,'' in \emph{Proceedings of the IEEE international
  conference on computer vision}, 2017, pp. 764--773.

\bibitem{carion2020end}
N.~Carion, F.~Massa, G.~Synnaeve, N.~Usunier, A.~Kirillov, and S.~Zagoruyko,
  ``End-to-end object detection with transformers,'' in \emph{European
  conference on computer vision}.\hskip 1em plus 0.5em minus 0.4em\relax
  Springer, 2020, pp. 213--229.

\bibitem{kuhn1955hungarian}
H.~W. Kuhn, ``The hungarian method for the assignment problem,'' \emph{Naval
  research logistics quarterly}, vol.~2, no. 1-2, pp. 83--97, 1955.

\bibitem{lin2017focal}
T.-Y. Lin, P.~Goyal, R.~Girshick, K.~He, and P.~Doll{\'a}r, ``Focal loss for
  dense object detection,'' in \emph{Proceedings of the IEEE international
  conference on computer vision}, 2017, pp. 2980--2988.

\bibitem{yang2019std}
Z.~Yang, Y.~Sun, S.~Liu, X.~Shen, and J.~Jia, ``Std: Sparse-to-dense 3d object
  detector for point cloud,'' in \emph{Proceedings of the IEEE/CVF
  International Conference on Computer Vision}, 2019, pp. 1951--1960.

\bibitem{yang20203dssd}
Z.~Yang, Y.~Sun, S.~Liu, and J.~Jia, ``3dssd: Point-based 3d single stage
  object detector,'' in \emph{Proceedings of the IEEE/CVF conference on
  computer vision and pattern recognition}, 2020, pp. 11\,040--11\,048.

\bibitem{he2020structure}
C.~He, H.~Zeng, J.~Huang, X.-S. Hua, and L.~Zhang, ``Structure aware
  single-stage 3d object detection from point cloud,'' in \emph{Proceedings of
  the IEEE/CVF Conference on Computer Vision and Pattern Recognition}, 2020,
  pp. 11\,873--11\,882.

\bibitem{li2021anchor}
J.~Li, H.~Dai, L.~Shao, and Y.~Ding, ``Anchor-free 3d single stage detector
  with mask-guided attention for point cloud,'' in \emph{Proceedings of the
  29th ACM International Conference on Multimedia}, 2021, pp. 553--562.

\bibitem{noh2021hvpr}
J.~Noh, S.~Lee, and B.~Ham, ``Hvpr: Hybrid voxel-point representation for
  single-stage 3d object detection,'' in \emph{Proceedings of the IEEE/CVF
  Conference on Computer Vision and Pattern Recognition}, 2021, pp.
  14\,605--14\,614.

\bibitem{zheng2021cia}
W.~Zheng, W.~Tang, S.~Chen, L.~Jiang, and C.-W. Fu, ``Cia-ssd: Confident
  iou-aware single-stage object detector from point cloud,'' in
  \emph{Proceedings of the AAAI conference on artificial intelligence},
  vol.~35, no.~4, 2021, pp. 3555--3562.

\bibitem{sheng2021improving}
H.~Sheng, S.~Cai, Y.~Liu, B.~Deng, J.~Huang, X.-S. Hua, and M.-J. Zhao,
  ``Improving 3d object detection with channel-wise transformer,'' in
  \emph{Proceedings of the IEEE/CVF International Conference on Computer
  Vision}, 2021, pp. 2743--2752.

\bibitem{chen2022sasa}
C.~Chen, Z.~Chen, J.~Zhang, and D.~Tao, ``Sasa: Semantics-augmented set
  abstraction for point-based 3d object detection,'' in \emph{AAAI Conference
  on Artificial Intelligence}, vol.~1, 2022.

\bibitem{he2020svga}
Q.~He, Z.~Wang, H.~Zeng, Y.~Zeng, S.~Liu, and B.~Zeng, ``Svga-net: Sparse
  voxel-graph attention network for 3d object detection from point clouds,''
  \emph{arXiv preprint arXiv:2006.04043}, 2020.

\bibitem{vora2020pointpainting}
S.~Vora, A.~H. Lang, B.~Helou, and O.~Beijbom, ``Pointpainting: Sequential
  fusion for 3d object detection,'' in \emph{Proceedings of the IEEE/CVF
  conference on computer vision and pattern recognition}, 2020, pp. 4604--4612.

\bibitem{pang2022fast}
S.~Pang, D.~Morris, and H.~Radha, ``Fast-clocs: Fast camera-lidar object
  candidates fusion for 3d object detection,'' in \emph{Proceedings of the
  IEEE/CVF Winter Conference on Applications of Computer Vision}, 2022, pp.
  187--196.

\bibitem{chen2022focal}
Y.~Chen, Y.~Li, X.~Zhang, J.~Sun, and J.~Jia, ``Focal sparse convolutional
  networks for 3d object detection,'' in \emph{Proceedings of the IEEE/CVF
  Conference on Computer Vision and Pattern Recognition}, 2022, pp. 5428--5437.

\bibitem{2204.00325}
Y.~Zhang, J.~Chen, and D.~Huang, ``Cat-det: Contrastively augmented transformer
  for multi-modal 3d object detection,'' 2022.

\bibitem{xie2020mlcvnet}
Q.~Xie, Y.-K. Lai, J.~Wu, Z.~Wang, Y.~Zhang, K.~Xu, and J.~Wang, ``Mlcvnet:
  Multi-level context votenet for 3d object detection,'' in \emph{Proceedings
  of the IEEE/CVF conference on computer vision and pattern recognition}, 2020,
  pp. 10\,447--10\,456.

\bibitem{zhang2020h3dnet}
Z.~Zhang, B.~Sun, H.~Yang, and Q.~Huang, ``H3dnet: 3d object detection using
  hybrid geometric primitives,'' in \emph{European Conference on Computer
  Vision}.\hskip 1em plus 0.5em minus 0.4em\relax Springer, 2020, pp. 311--329.

\bibitem{chen2020hierarchical}
J.~Chen, B.~Lei, Q.~Song, H.~Ying, D.~Z. Chen, and J.~Wu, ``A hierarchical
  graph network for 3d object detection on point clouds,'' in \emph{Proceedings
  of the IEEE/CVF conference on computer vision and pattern recognition}, 2020,
  pp. 392--401.

\bibitem{xie2021vote}
Q.~Xie, Y.-K. Lai, J.~Wu, Z.~Wang, Y.~Zhang, K.~Xu, and J.~Wang, ``Vote-based
  3d object detection with context modeling and sob-3dnms,''
  \emph{International Journal of Computer Vision}, vol. 129, no.~6, pp.
  1857--1874, 2021.

\bibitem{liu2021group}
Z.~Liu, Z.~Zhang, Y.~Cao, H.~Hu, and X.~Tong, ``Group-free 3d object detection
  via transformers,'' in \emph{Proceedings of the IEEE/CVF International
  Conference on Computer Vision}, 2021, pp. 2949--2958.

\bibitem{lahoud20172d}
J.~Lahoud and B.~Ghanem, ``2d-driven 3d object detection in rgb-d images,'' in
  \emph{Proceedings of the IEEE international conference on computer vision},
  2017, pp. 4622--4630.

\bibitem{ren2016three}
Z.~Ren and E.~B. Sudderth, ``Three-dimensional object detection and layout
  prediction using clouds of oriented gradients,'' in \emph{Proceedings of the
  IEEE conference on computer vision and pattern recognition}, 2016, pp.
  1525--1533.

\bibitem{geiger2012we}
A.~Geiger, P.~Lenz, and R.~Urtasun, ``Are we ready for autonomous driving? the
  kitti vision benchmark suite,'' in \emph{2012 IEEE conference on computer
  vision and pattern recognition}.\hskip 1em plus 0.5em minus 0.4em\relax IEEE,
  2012, pp. 3354--3361.

\bibitem{song2015sun}
S.~Song, S.~P. Lichtenberg, and J.~Xiao, ``Sun rgb-d: A rgb-d scene
  understanding benchmark suite,'' in \emph{Proceedings of the IEEE conference
  on computer vision and pattern recognition}, 2015, pp. 567--576.

\end{thebibliography}

\vfill

\end{document}